\documentclass[lettersize,journal]{IEEEtran}
\usepackage{amsmath,amsfonts}
\usepackage{array}
\usepackage{textcomp}
\usepackage{stfloats}
\usepackage{url}
\usepackage{verbatim}
\usepackage{graphicx}
\usepackage{cite}
\usepackage{xcolor}
\usepackage{subcaption}
\usepackage{algorithm}
\usepackage{algpseudocode}
\usepackage{dsfont}
\usepackage{hyperref}
\hypersetup{
  colorlinks   = true,              
  urlcolor     = black,              
  linkcolor    = black,              
  citecolor    = black                
}
\usepackage{booktabs}
\usepackage{adjustbox}
\usepackage[english]{babel}
\usepackage{xspace}

\addto\extrasenglish{  
    
}

\addto\extrasenglish{  
    
}
\addto\extrasenglish{  
    
}
\addto\extrasenglish{  
    
}
\addto\extrasenglish{  
    
}

\newcommand{\pcgnn}{PCGNN\xspace}
\newcommand{\mcc}{\texttt{MCHAMR}\xspace}
\newcommand{\minecraft}{Minecraft\xspace}

\newcommand{\newrev}[1]{#1}

\newcommand{\newtxt}[1]{#1}

\newcommand{\appendixLetterExt}{A\xspace}
\newcommand{\appendixLetterRepr}{B\xspace}
\newcommand{\appendixLetterExpSetup}{B\xspace}
\newcommand{\appendixLetterHyperParams}{C\xspace}
\newcommand{\appendixLetterCoalesce}{D\xspace}
\newcommand{\appendixLetterQuantNov}{E\xspace}

\newcommand{\mytitle}{Hierarchically Composing Level Generators for the Creation of Complex Structures}
\newcommand{\myauthors}{Michael Beukman, Manuel Fokam, Marcel Kruger, Guy Axelrod, Muhammad Nasir, \\ Branden Ingram, Benjamin Rosman, Steven James\thanks{School of Computer Science and Applied Mathematics, University of the Witwatersrand.}}

\newcommand{\ghurl}{https://github.com/Michael-Beukman/MCHAMR}

\newcommand\EatDot[1]{}
\makeatletter
\newcommand*{\centerfloat}{%
  \parindent \z@
  \leftskip \z@ \@plus 1fil \@minus \textwidth
  \rightskip\leftskip
  \parfillskip \z@skip}
\makeatother

\newcommand{\refappendix}[1]{\hyperref[#1]{Appendix}}
\newcommand{\refappendixExt}[1]{\hyperref[sec:appdx:extensions]{Appendix \appendixLetterExt}}
\newcommand{\refappendixRepr}[1]{\hyperref[sec:appdx:repr]{Appendix \appendixLetterRepr}}
\newcommand{\refappendixExpSetup}[1]{\hyperref[sec:appdx:expsetup]{Appendix \appendixLetterExpSetup}}
\newcommand{\refappendixHyperParams}[1]{\hyperref[sec:appdx:hyperparams]{Appendix \appendixLetterHyperParams}}
\newcommand{\refappendixCoalesce}[1]{\hyperref[sec:appdx:coalesce]{Appendix \appendixLetterCoalesce}}
\newcommand{\refappendixQuantNov}[1]{\hyperref[sec:appdx:novresults]{Appendix \appendixLetterQuantNov}}

\newcommand{\refappendixExtsAndHyperparams}[1]{Appendices \hyperref[sec:appdx:extensions]{\appendixLetterExt} and \hyperref[sec:appdx:hyperparams]{\appendixLetterHyperParams}}

\makeatletter
\newcommand\Autoref[1]{\@first@ref#1,@}
\def\@throw@dot#1.#2@{#1}
\def\@set@refname#1{
    \edef\@tmp{\getrefbykeydefault{#1}{anchor}{}}%
    \xdef\@tmp{\expandafter\@throw@dot\@tmp.@}%
    \ltx@IfUndefined{\@tmp autorefnameplural}%
         {\def\@refname{\@nameuse{\@tmp autorefname}s}}%
         {\def\@refname{\@nameuse{\@tmp autorefnameplural}}}%
}
\def\@first@ref#1,#2{%
  \ifx#2@\autoref{#1}\let\@nextref\@gobble
  \else%
    \@set@refname{#1}
    \@refname~\ref{#1}
    \let\@nextref\@next@ref
  \fi%
  \@nextref#2%
}
\def\@next@ref#1,#2{%
   \ifx#2@ and~\ref{#1}\let\@nextref\@gobble
   \else, \ref{#1}
   \fi%
   \@nextref#2%
}

\makeatother

\makeatletter

\newlength{\phaserulewidth}
\newcommand{\setphaserulewidth}{\setlength{\phaserulewidth}}

\algrenewcommand\algorithmicindent{1em}%

\makeatother

\setphaserulewidth{.7pt}

\usepackage[numbers]{natbib}

\begin{document}

\title{\mytitle}

\author{\myauthors}

\maketitle
\begin{abstract}
  Procedural content generation (PCG) is a growing field, with numerous applications in the video game industry and great potential to help create better games at a fraction of the cost of manual creation. 
  However, much of the work in PCG is focused on generating relatively straightforward levels in simple games, as it is challenging to design an optimisable objective function for complex settings.
  This limits the applicability of PCG to more complex and modern titles, hindering its adoption in industry. Our work aims to address this limitation by introducing a compositional level generation method that recursively composes simple low-level generators to construct large and complex creations. This approach allows for easily-optimisable objectives and the ability to design a complex structure in an interpretable way by referencing lower-level components. We empirically demonstrate that our method outperforms a non-compositional baseline by more accurately satisfying a designer's functional requirements in several tasks. Finally, we provide a qualitative showcase (in \minecraft) illustrating the large and complex, but still coherent, structures that were generated using simple base generators.
\end{abstract}

\section{Introduction}
\label{chap:intro}
Procedural content generation (PCG) is a research field focused on automatically generating game content, such as levels, maps and music~\citep{togelius2011Searchbased}. PCG has several benefits: it is often cheaper than manually designing content~\citep{hendrikx2013Procedural}, and it allows for significantly more content to be generated than would otherwise be possible through human creation~\citep{smith2017Procedural,korn2017Procedural}. 
PCG has recently garnered more attention~\citep{brewer2017computerized}, leading to impressive results in numerous games~\citep{volz2018Evolving,gisslen2021Adversarial,earle2021Illuminating}.

However, much of the research done in the field of PCG focuses on simple, 2D tilemap games, such as \textit{Super Mario Bros}, \textit{The Legend of Zelda}, and maze games~\citep{beukman2022Procedural,khalifa2020Pcgrl,earle2021Illuminating,ferreira2014Multipopulation,shaker2011Mario}. While these are useful testbeds, they are often very simple, and do not exhibit complex structures. This focus on simplicity makes it non-trivial to generalise these methods to more complex and modern games. Furthermore, few methods focus on, or excel at, generating large and complex levels, which is a necessary step for PCG to become more mainstream~\citep{togelius2013Procedural}.

This has led to recent research that applies PCG to more complex titles~\citep{grbic2020Evocraft,salge2018Generative,jiang2022Learning,barthet2022openended}, particularly \minecraft. \minecraft is a 3D, discrete voxel-based game, making it amenable to grid-based PCG methods originally developed for 2D tilemaps. It is also more complex than these aforementioned games and thus provides a useful domain to bridge the gap between traditional and modern games.

Despite progress in the field, there are still several shortcomings to current methods.
\newrev{Some rely on a developer manually designing level components~\citep{salge2021Settlement}}, which can be costly~\citep{hendrikx2013Procedural} and difficult to transfer to new games~\citep{khalifa2020Pcgrl}. Other approaches generate levels that lack core functional requirements, such as being playable~\citep{grbic2020Evocraft}. Lastly, some methods generate relatively small and simple structures~\citep{jiang2022Learning,barthet2022openended}, limiting their applicability to complex games.

Our work aims to address these limitations by proposing an approach that decomposes the problem of generating a complex and large-scale level into smaller, but manageable pieces. For instance, it is simple to generate an abstract, high-level layout of a town (by considering houses, roads and gardens as the individual building blocks) while ensuring all houses are reachable via roads. Similarly, it is relatively straightforward to learn to generate a single house, garden, or farm. Composed together, however, these simple generators can easily generate a fully-fledged, and functional, town.

In our method, multiple generators are hierarchically composed to generate large and complex structures. This is inspired by \textit{hierarchical reinforcement learning}~\citep{pateria2021Hierarchical}, where a complex task is broken down into smaller components and an agent learns how to optimally perform each subcomponent. We break the construction of a high-level structure (e.g., a town) into simple subtasks (e.g., houses, gardens, etc.) and train \newrev{independent} agents to generate these individual components.
\newrev{Furthermore, instead of manually designing level elements, our method allows a designer to specify an objective function, i.e., what constitutes a ``good'' component.}

This approach has three main benefits: (1) \newrev{due to the independence, } users can modify the creation of each component (at different levels of abstraction) in isolation; (2) the combinatorial explosion in the number of levels that can be generated~\citep{tasse2020Boolean}; and (3) the ability to specify the desired structure by referencing subcomponents, allowing easier generation of complex levels compared to normal, non-hierarchical methods.

We extend \pcgnn~\citep{beukman2022Procedural}, a recent method which uses neuroevolution and novelty search to evolve diverse level generators, by hierarchically composing these \newrev{independent} generators to create large and complex structures \newrev{in tile-based games}. We demonstrate that using composition makes it significantly easier to generate complex structures compared to using a single, non-compositional generator, an effect which becomes more pronounced as we increase the complexity of the low-level structures. Furthermore, we can automatically generate large-scale towns and cities by specifying a few, simple objectives.\footnote{Code is available at \url{\ghurl}}

\section{Background}
\label{chapter:background}
\label{sec:bg}

\subsection{Hierarchical and Compositional Reinforcement Learning}
Reinforcement learning (RL)~\citep{rlbookold} is a field focused on solving sequential decision-making problems, where an agent interacts with an environment (e.g., by placing specific tiles in a tilemap level) and receives a scalar reward signal indicating how good an action was (e.g., obtaining a high reward when removing a tile that creates a path in a maze~\citep{khalifa2020Pcgrl}).

However, \textit{long-horizon tasks} require a long sequence of actions to solve (e.g., generating a level by sequentially placing many tiles) and are often challenging for standard RL methods~\citep{pateria2021Hierarchical,nachum2019Does}. 
Hierarchical RL addresses this by decomposing complex tasks into (1) smaller subtasks and a (2) top-level policy that selects which subtask to perform~\citep{pateria2021Hierarchical,hengst2012Hierarchical}. 
In compositional RL, agents compose low-level skills to perform more complex behaviours, allowing tasks to be specified in a more understandable way by referencing these skills~\citep{tasse2020Boolean}.

\subsection{Evolutionary Algorithms}
Many PCG methods~\citep{togelius2011Searchbased,ferreira2014Multipopulation,earle2021Illuminating} use aspects of evolutionary computing, a subclass of optimisation algorithms that attempt to mimic natural selection~\citep{goldberg1989genetic}.
Genetic algorithms (GAs) usually consist of a collection (or population) of individuals, each possessing a \textit{genome} which describes the individual (e.g., an integer vector representing the height of a platform~\citep{ferreira2014Multipopulation}). These genomes are mapped to \textit{phenotypes}, which can be thought of as the result or instantiation of this individual in the actual problem domain (e.g., a level generated using the given height information).
To create the next generation, the current population's high-performing individuals (determined by the \textit{fitness function}) are merged through \textit{crossover} to produce children, which are then subjected to slight mutations.

\subsubsection{NeuroEvolution of Augmenting Topologies (NEAT)}
NEAT~\citep{neat} is a genetic algorithm that evolves the weights and structure of neural networks. NEAT starts with a population of simple networks with no hidden layers, which gradually increase in complexity through evolution, to better solve the problem under consideration.

\subsubsection{Novelty Search}
Novelty search~\citep{novelty_search} rewards individuals based on their novelty compared to the population, instead of their objective performance (as in traditional genetic algorithms). This results in improved exploration and can help to solve hard or deceptive\footnote{Deceptive problems are those where directly maximising fitness can lead to being stuck in local minima~\citep{whitley1991Fundamental,novelty_search}.} problems.

\subsection{\pcgnn}
\pcgnn~\citep{beukman2022Procedural} is a method that uses NEAT~\citep{neat} and novelty search~\citep{novelty_search} to evolve level generators. The main goal of this work is to learn reusable level generators that can quickly generate multiple different levels, as opposed to searching for single levels each time one is desired.  Each individual in this population is a neural network, which can be used to generate multiple levels. To obtain a level from a network, an initial, random, tilemap is created. Then, for each tile, the neural network receives as input the surrounding tiles and outputs the current center tile. This process is repeated, sequentially, for each tile. \pcgnn generally uses multiple fitness functions: (1) the normal novelty search objective, to incentivise exploration; (2) intra-generator novelty, which incentivises one neural network to generate multiple diverse levels; and (3) one or more \textit{feasibility} fitness functions, which describe the feasibility criteria for the specific game (e.g., solvability in a maze). 
\pcgnn generates levels quickly and can generalise to unseen level sizes, allowing the generation of arbitrarily-sized levels without retraining.

\section{Related Work}
\subsection{General PCG}
Although there are numerous approaches to PCG, many focus on generating simple levels, and are therefore difficult to generalise to more complex settings, such as modern games. 
Search-based techniques, such as evolutionary algorithms, are commonly used to generate levels that maximise a specific objective function~\citep{togelius2011Searchbased,beukman2022Procedural,earle2021Illuminating,ferreira2014Multipopulation,barthet2022openended}.
Similarly, the recent paradigm of Procedural Content Generation via Reinforcement Learning generates levels using RL, training an agent policy to sequentially place tiles to maximise some reward~\citep{khalifa2020Pcgrl,earle2021Learning,jiang2022Learning}. 
However, it is challenging to design a monolithic objective or reward function that incentivises the generation of complex and functional structures while being optimisable~\citep{togelius2013Procedural}. 
Other methods rely on a dataset of existing content (which may not be available for many games), applying machine learning to generate novel content~\citep{summerville2017Procedural,liu2021Deep,shu2021Experiencedriven}. 

\subsubsection{Generating Complex Levels}

While many approaches focus on simple 2D tile-based games, such as \textit{Super Mario Bros} and \textit{The Legend of Zelda}, other work rather aims to generate more complex content. 
One notable example is the \textit{Generative Design in Minecraft (GDMC)} competition, where the task is to generate an entire settlement in \minecraft~\citep{salge2018Generative, salge2021Settlement}. 
This competition aims to improve the generation of holistic and complex constructions (e.g., an entire, coherent town instead of just a single building) while taking the external environment into account.
Many submissions to this competition, however, focused on more hand-designed, rule-based~\citep{barthet2022openended} methods specifically designed for settlement generation, which may require significant effort to generalise to other scenarios.
This has led to work that focuses on automatically generating lower-level structures one would find in a settlement, such as buildings. For instance, \citet{green2019Organic} use constrained growth algorithms to separately generate the building and floor plan, which can be useful when customisation is desired.
\citet{barthet2022openended} use DeLeNoX~\citep{liapis2021Transforming}, to evolve diverse building generators for \minecraft using an autoencoder-based novelty score. 
Here, the focus is on generating creative buildings, rather than generating complex and functional structures or how to combine these structures into an actual settlement. 

\subsubsection{Adversarial and Open-Ended Approaches}

Recently, PCG has also been used to generate a large number of diverse levels on which machine learning agents can be trained, leading to these agents becoming more robust~\citep{justesen2018illuminating,cobbe2019quantifying}.
Level generators and game-playing agents can often work in tandem, where the generator generates levels that challenge the agent. As the agent improves, the level generator must also improve to generate more complex and harder levels~\citep{bontrager2020Learning,gisslen2021Adversarial}. 

Most of these methods, however, focus on simple games (such as mazes~\citep{holder2022Evolving} or 2D terrains~\citep{wang2019POET}), making it unclear how to generalise them to more complex games. 
Furthermore, the focus is often on obtaining robust and high-performing artificial agents~\citep{holder2022Evolving} instead of generating video game levels to be played and enjoyed by humans. Prioritising the former could lead to overly difficult levels.
These levels may also have different characteristics compared to human-designed ones~\citep{dharna2020Cogeneration}, complicating their use as a substitute for manual content creation.
Lastly, treating the generation process as one large problem may limit the ability of the generator to create levels with enough complexity to challenge the machine learning agent.
Instead, having multiple generators that solve simpler problems at different levels of abstraction could enable the generation of more complex and larger-scale structures~\citep{snodgrass2015Hierarchical}.

\subsection{Compositional PCG}
Much work has also been done on combining different PCG methods to generate more complex pieces of content, or even entire games~\citep{liapis2019Orchestrating}. For instance, \citet{togelius2012Compositional} combine an Answer Set Programming (ASP)-based approach~\citep{smith2011Answer} with a genetic algorithm that searches over variables for the ASP. 
\citet{liapis2019Orchestrating} advocate for game design orchestration, where different content generators are combined to generate full, coherent, games. One benefit of this compositional approach is allowing algorithms to specialise in (and thus excel at generating) specific pieces of content. There are still numerous open problems in this field, however, such as how to best coordinate automated generators and human designers.

Hierarchy can also be used to improve the generation speed of PCG methods. For example, \citet{smith2014Logical} find that using ASPs to generate large levels is infeasibly slow but that using hierarchy can lead to much faster generation. They first generate the high-level structure of a level and subsequently fill in the details within this fixed structure. However, this approach uses hand-designed high-level structures, which may not be easily generalisable to new games~\citep{togelius2013Procedural}.

\citet{snodgrass2015Hierarchical} also leverage hierarchy to generate structures at different scales. From a set of example levels, they extract a set of high-level and low-level patterns and fit two multi-dimensional Markov chains to this data. To generate a level, then, the high-level model generates the structure while the low-level generator fills in the details. This method results in improved generation compared to a non-hierarchical approach, but requires example levels as training data, which may not always be available~\citep{summerville2017Procedural}.

Hierarchy can also be used to separately construct different aspects of a level. 
For instance, \citet{dormans2010Adventures} first generates the ``mission'': the high-level task that the player must complete. The physical layout of the level is then generated based on the mission.
Similarly, in \textit{Unexplored}~\citep{unexplored}, a graph representing the level is created by applying a set of replacement rules to create, remove and modify nodes to obtain certain properties. 
This graph is then mapped onto a grid, and each node is transformed into a room~\citep{unexploredWebsiteExplain}. Although this approach is hierarchical, each of the 5000 replacement rules is hand-designed, hindering application to other settings.

Another use of composition is to increase the resolution of high-level sketches provided by a designer. For instance, \citet{liapis2013SentientWorld} allow a designer to draw a high-level, low-resolution sketch of a level. Then, using techniques such as genetic algorithms, the PCG system replaces each high-level tile with a more detailed and fleshed-out component. While this can be used to obtain a high-resolution version of a designer's sketch, it may be less suitable when a complex structure consisting of several specific components is desired.
Instead of hand-designing the high-level sketch, \citet{liapis2017Multi} allows the designer to specify a fitness function that is used to evolve it. The sketch is then parsed into a graph indicating which segments are connected, thereby defining constraints for each segment (such as the number and location of doors). Using these constraints, each segment is evolved and placed in the high-level sketch to form a complete level. While this work is promising, it focuses on dungeon levels, without a clear way to apply the same technique when generating other types of levels. Furthermore, this approach directly evolves the sketch and the low-level segments, meaning that evolution occurs at generation time, which may be prohibitively slow~\citep{beukman2022Procedural}.

\citet{dormans2013Combinatorial} take a different perspective: their two-step process first generates a large amount of diverse content, and the second step reorganises this content into an actual level. 
While this simplifies the generation step, the content is reorganised without any hierarchical elements.

Other methods create levels by generating and combining different \textit{layers}\newrev{, effectively composing together different generators. However, these methods generally focus on iteratively furnishing a level, as opposed to hierarchically generating a complex layout.}
For instance, \citet{green2019Two} first generate the structure of a dungeon level---tiles such as walls and the floor. Then, a different method furnishes this structure by placing items such as treasure and enemies.
Also following a similar layered approach, \citet{wu2021Procedural} generate natural levels. They break the level down into several layers such as ocean, land, and forests, with each layer being generated by a cellular automaton (CA). 
This approach is a promising way to generate landscapes, but the handcrafted rules could be difficult to generalise to other domains. Furthermore, although CAs can generate landscapes, they may be less well-suited to generating structures that satisfy certain complex constraints\newrev{---for instance, cities where houses must be reachable by roads}.
\textit{Dwarf Fortress} \citep{dwarffortress} also uses multiple layers; here, the world, history and several other aspects evolve over time in a rule-based simulation. This process begins by first generating the world and progressively adding more specialised designed systems.
The world is created by generating an elevation map using a randomised fractal, followed by multiple layers of maps including temperature, rainfall and vegetation. Once the world is created, the simulation of civilisation begins where settlements are built, trade routes are formed, and wars are fought. The simulation stops at a designated point in time, and the player starts the game in a unique living world.

\begin{figure*}[ht]
  \captionsetup{width=.24\linewidth}
  \centering
  \begin{subfigure}[t]{0.245\linewidth}
    \resizebox{1\linewidth}{1\linewidth}{\includegraphics[width=\linewidth]{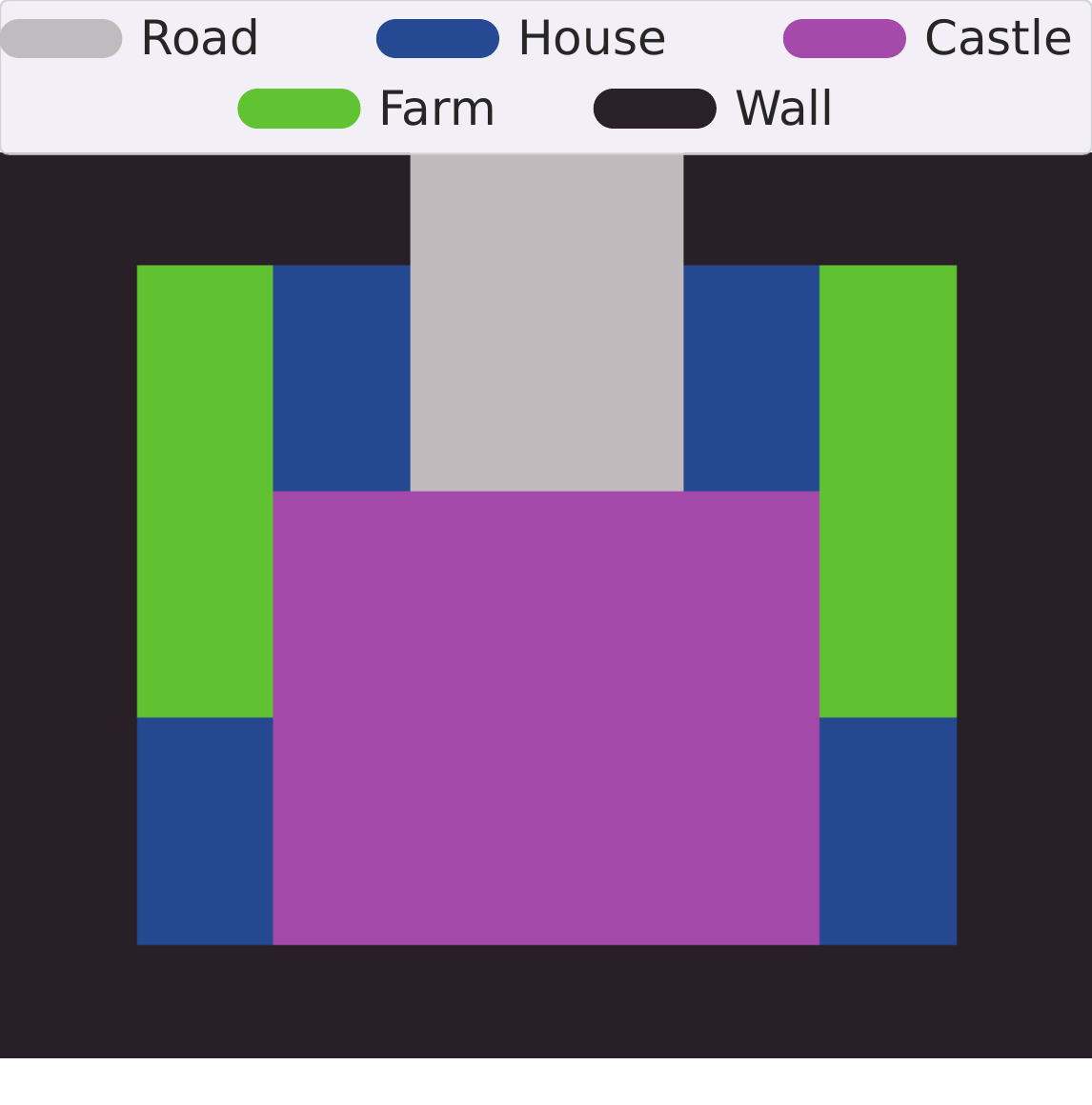}}
    \caption{High-level view of a town.}
    \label{fig:mcc:we:1-town}
  \end{subfigure}\hfill
  \begin{subfigure}[t]{0.245\linewidth}
    \resizebox{1\linewidth}{1\linewidth}{\includegraphics[width=\linewidth]{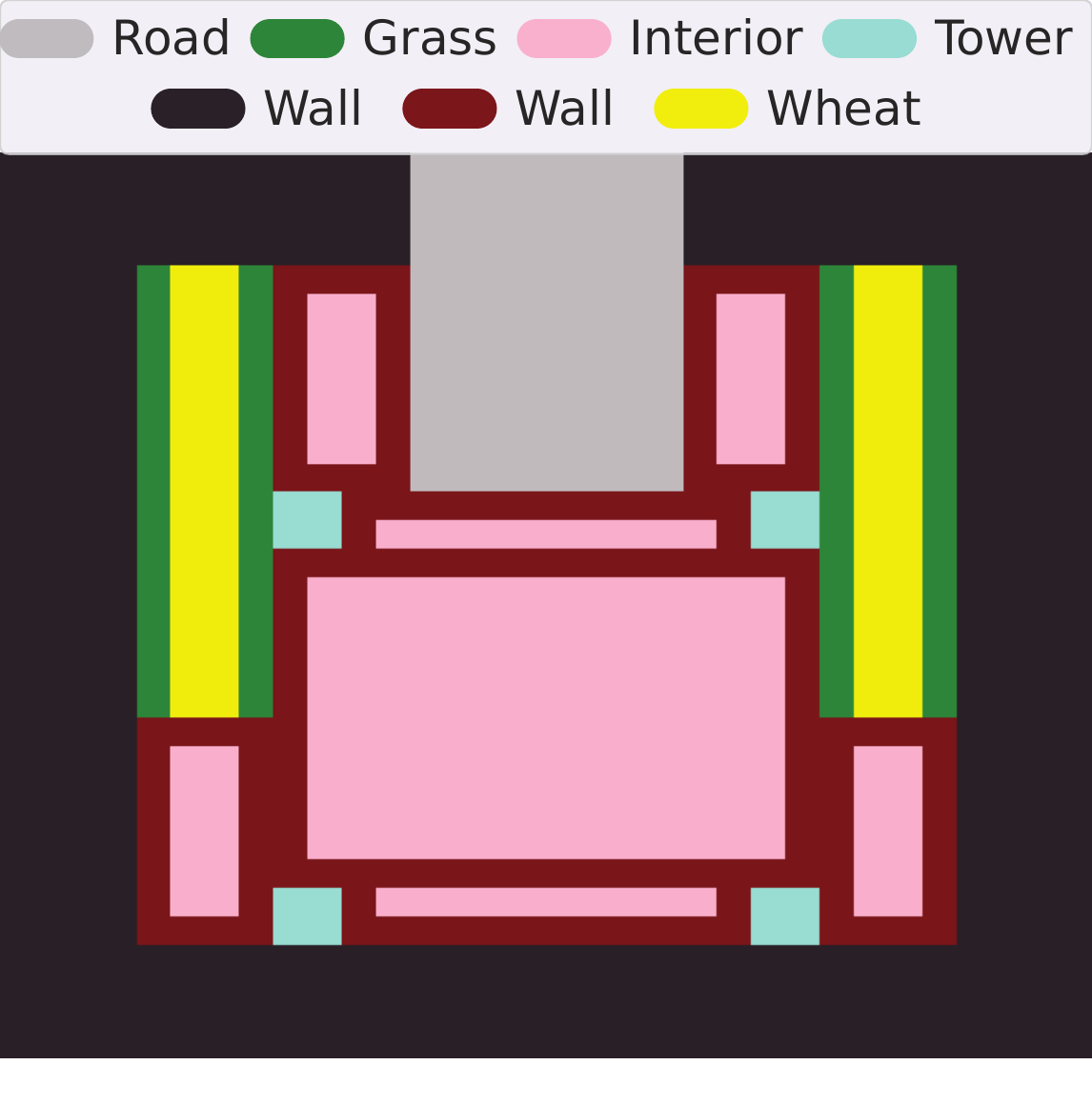}}
    \caption{The town after composition.}
    \label{fig:mcc:we:3-composed}
  \end{subfigure}\hfill
  \begin{subfigure}[t]{0.245\linewidth}
    \resizebox{1\linewidth}{1\linewidth}{\includegraphics[width=\linewidth]{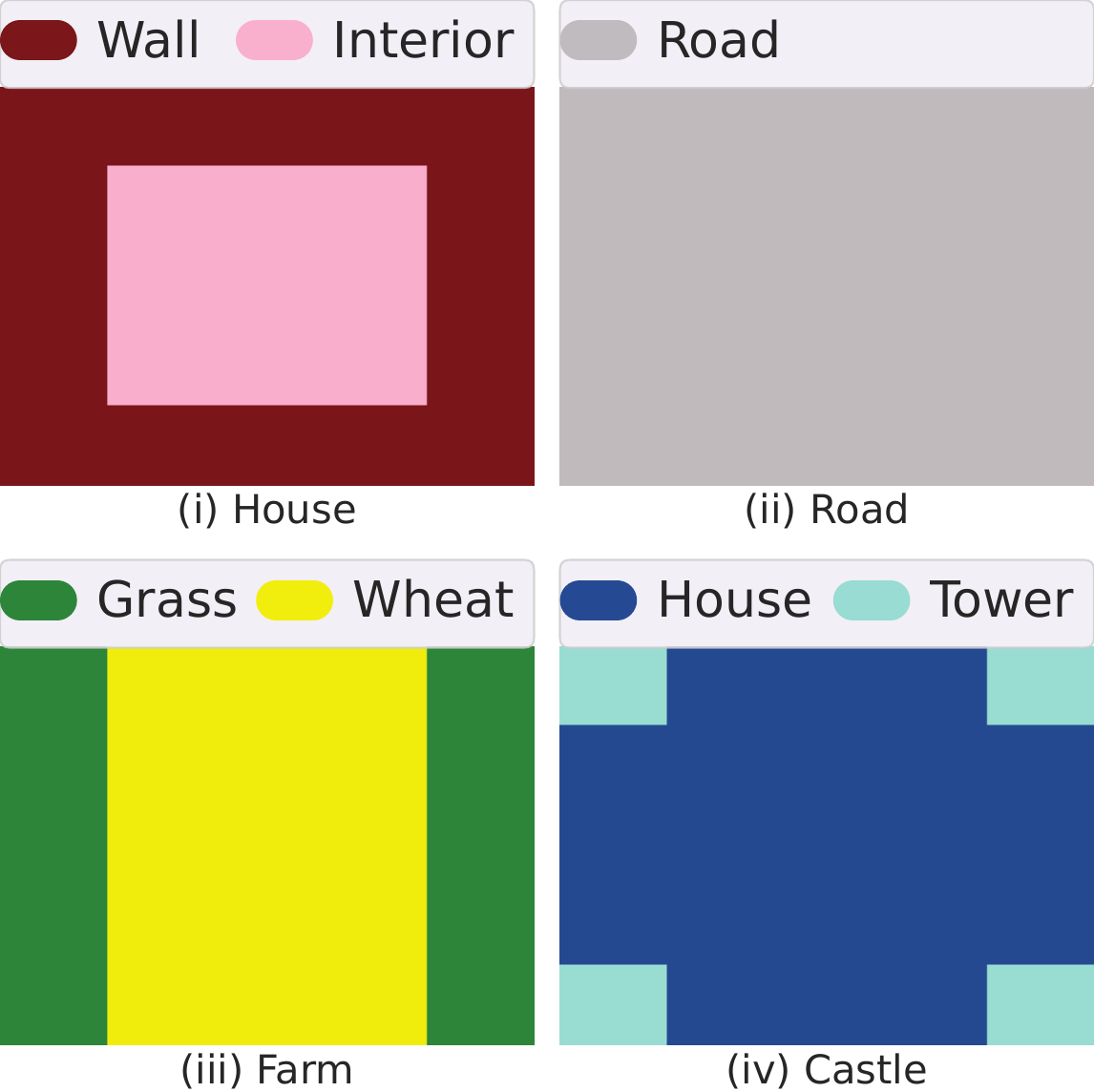}}
    \caption{Base components.}
    \label{fig:mcc:we:2-components}
  \end{subfigure}\hfill
  \begin{subfigure}[t]{0.245\linewidth}
    \resizebox{1\linewidth}{1\linewidth}{\includegraphics[width=1\linewidth]{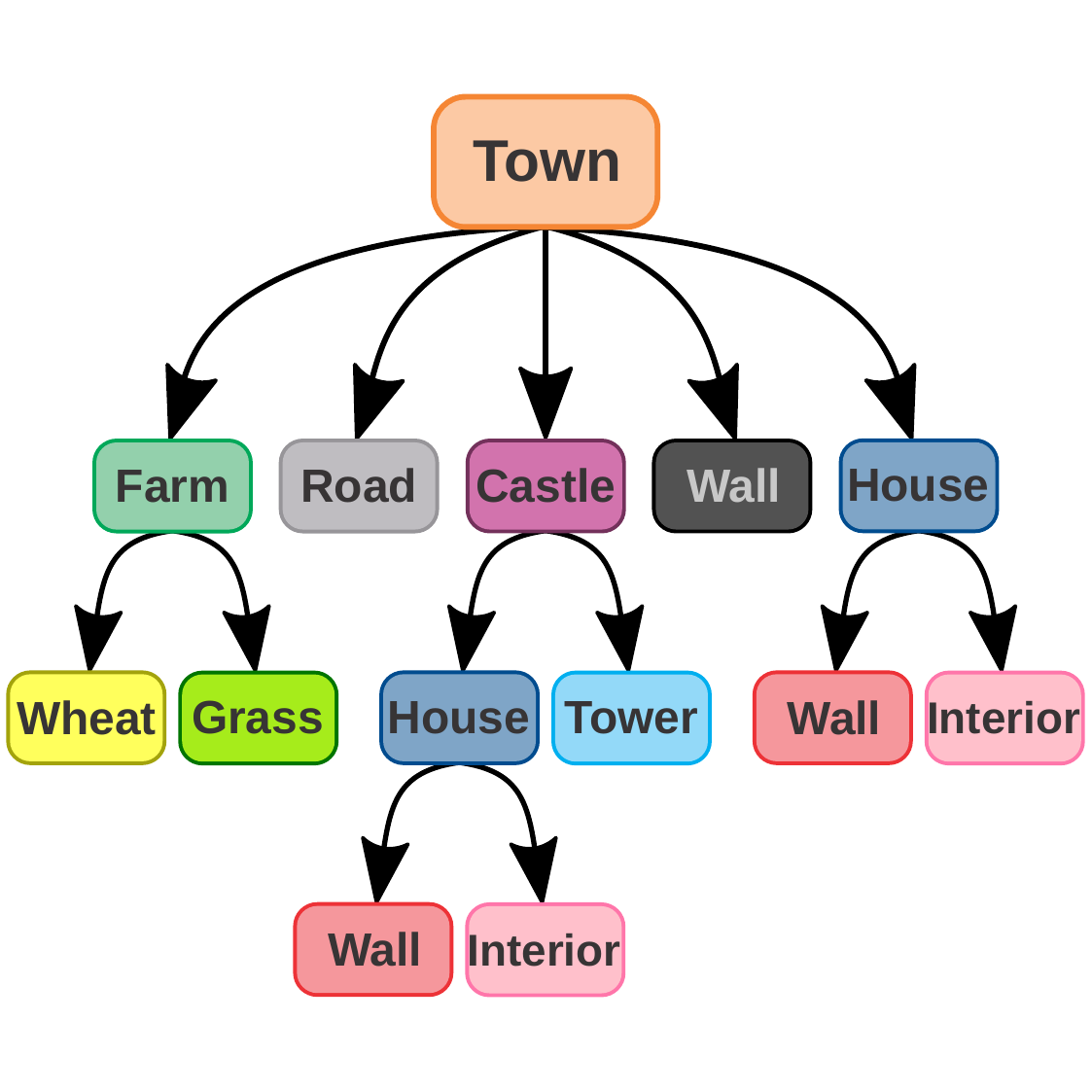}}
    \caption{The tree structure for (b).}
    \label{fig:mcc:high_level_tree}
  \end{subfigure}
  \captionsetup{width=1\linewidth}
  \caption{An example of \mcc, with the (a) high-level and (b) composed towns and (c) the base components. The tree structure defining how components are combined is shown in (d), with leaf nodes corresponding to low-level tiles.}
\end{figure*}
\section{Hierarchically Composing Level Generators}\label{sec:method:mcc}
Here we describe our approach, \textbf{M}ulti-level \textbf{C}omposition of \textbf{H}ierarchical and \textbf{A}daptive \textbf{M}odels via \textbf{R}ecursion (\mcc).

To illustrate our method, we first consider the example of generating a simple medieval town. At a high level, we can represent the town as a 2D tilemap, shown in \autoref{fig:mcc:we:1-town}, with houses, a large castle and roads. Each of these components, however, can be decomposed into simpler components; for instance, a house that is a single tile in \autoref{fig:mcc:we:1-town} actually consists of external walls and an empty interior, as shown in \autoref{fig:mcc:we:2-components}. Using our composition algorithm (described below), we can use the definitions of these smaller structures together with the high-level map to obtain a fully-fledged and detailed town, shown in \autoref{fig:mcc:we:3-composed}. Furthermore, the user has control over each component. Should they wish to make a similar town but would prefer to replace or modify a specific structure, this is possible without altering any of the other components.

To implement this method in an automatic way, we require the concept of a \textit{generator} $g \in G$. This object can generate a tilemap $W$, which is represented as a 2- or 3-dimensional matrix, where each element is taken from some tile set $T$. Each generator optionally has a \textit{tile mapping} $M_g: T \to G$ that controls which subgenerators are used when this generator places a specific tile. For instance, in the above example, the town generator's mapping specifies that the ``house'' tile actually maps to the generator shown in the top left pane of \autoref{fig:mcc:we:2-components}. In essence, we construct a tree (\autoref{fig:mcc:high_level_tree} shows the tree for the previous example), where each node is a level generator, with parent nodes generating placeholders to be filled by their children and leaf nodes producing the lowest-level, non-abstract, structures. To generate the level, we start at the root and traverse through the tree. 

Finally, for each generator $g$, we also have $S_g$, which indicates the size of the subtiles of this generator, i.e., the size at which the next level of the hierarchy will be generated. For example, if our town generator $g_{town}$ produces an abstract tilemap of size $10\times 10$, and has an associated $S_{g_{town}}$ of $3\times 3$, then the final level will have size $30\times 30$. Each tile in the abstract map corresponds to a $3\times 3$ block of low-level tiles.

\begin{figure*}
    \centering
    \includegraphics[width=1\linewidth]{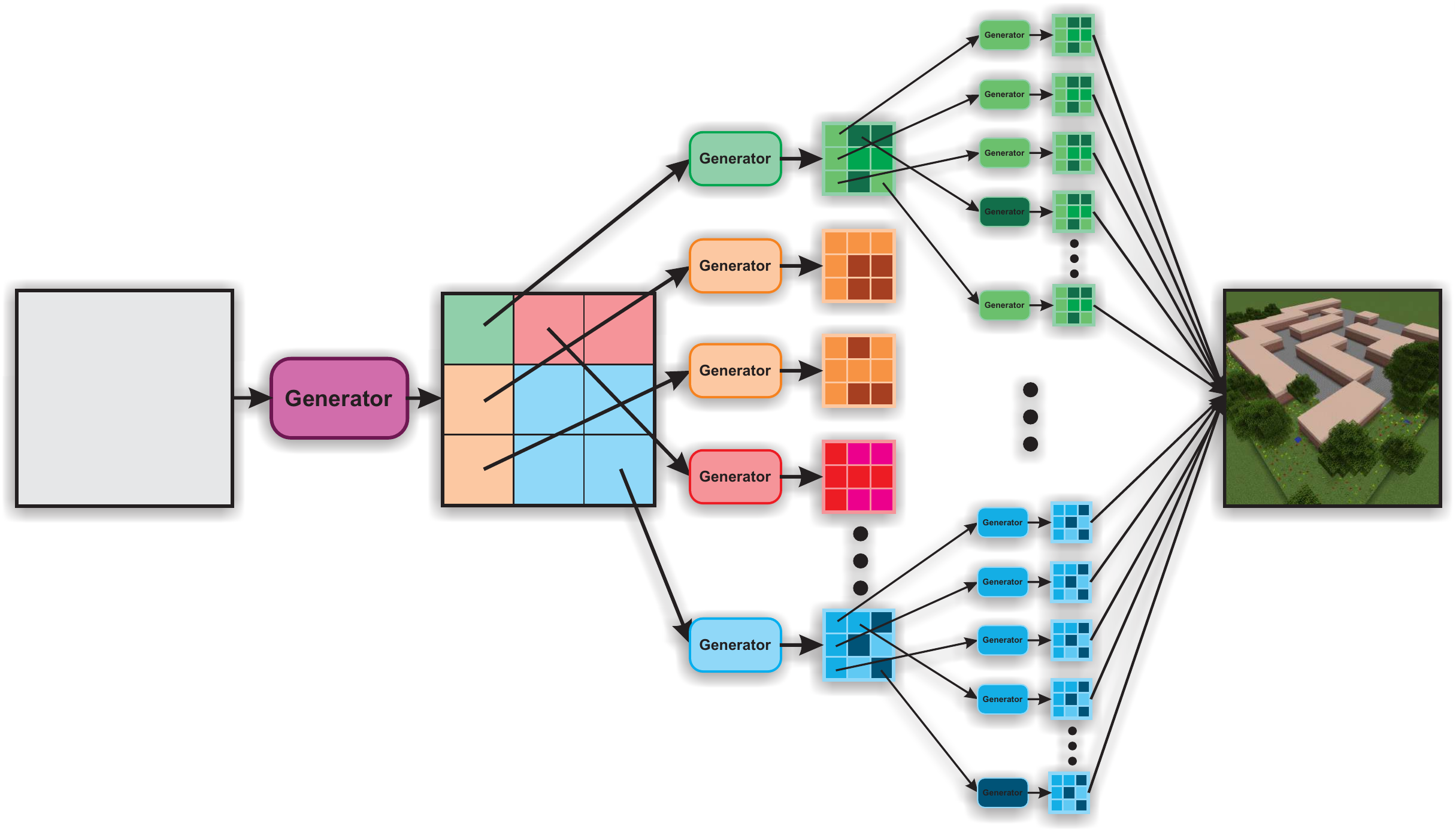}
    \caption{A high-level illustration of \mcc. We begin with a top-level generator, which generates an abstract tilemap. Each tile in this map is an instruction to use a lower-level generator and to place its output at that location. Each of these generators can be further composed of other generators. In the end, we obtain a fully-fledged level.}
    \label{fig:mcc:high_level_tldr}
\end{figure*}

Given a set of generators, our algorithm (detailed in \autoref{alg:mcc}) works as follows: We start with the top-level generator, which generates a tilemap (line 4 in \autoref{alg:mcc}). In standard PCG, the procedure would terminate here. We instead go further and interpret each tile as an instruction to use another generator and place its output at that location (lines 14-16). Each of these lower-level generators can also be composed of generators. If at any point, though, the generator does not contain a \textit{tile mapping} $M_g$, we end the recursion (lines 5-6). 
The overall generation process is illustrated in \autoref{fig:mcc:high_level_tldr}.

\begin{figure}
  \centering
  \begin{subfigure}{0.4\linewidth}
      \includegraphics[width=1\linewidth]{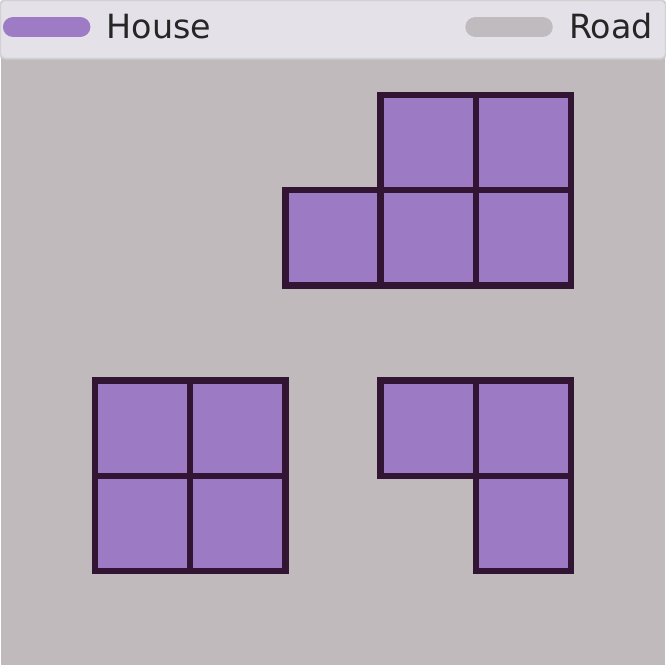}
      \caption{}
  \end{subfigure}
  \begin{subfigure}{0.4\linewidth}
      \includegraphics[width=1\linewidth]{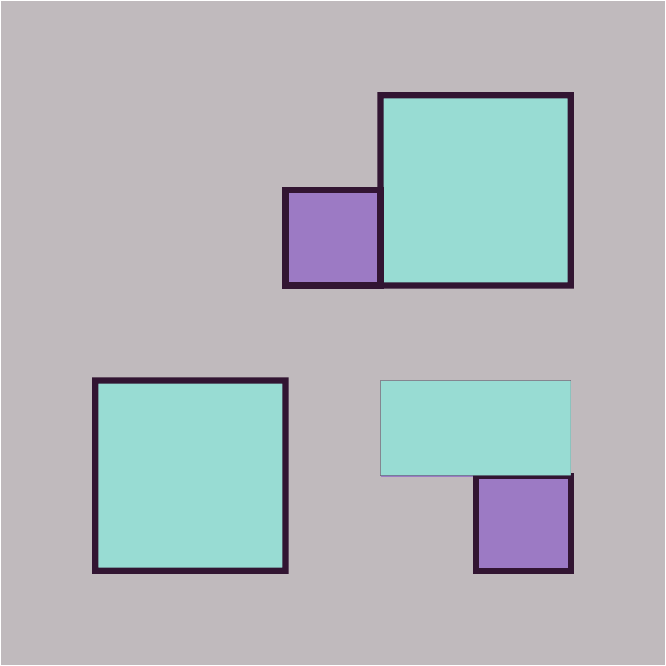}
      \caption{}
  \end{subfigure}
  \caption{(a) A level, and (b) the result after coalescing connected tiles. Cyan tiles were coalesced into one larger tile.}
  \label{fig:mcc:coalesce}
\end{figure}

This method tends to produce blocky structures since each lower-level tile is of the same size. To overcome this, we combine connected tiles of the same type into a single, larger, ``logical'' tile (line 12). Then, at generation time, we generate one large structure of this size, instead of many small ones.
To achieve this, we find and merge contiguous rectangles by attempting to expand the dimensions of each tile one at a time until we can no longer do so. We repeat this process at every level of the hierarchy, coalescing connected tiles before recursively calling the subgenerators.
Coalescing not only improves the cohesiveness of the generated structure but also gives the top-level generator more control over the scale of the lower-level components. \autoref{fig:mcc:coalesce} illustrates this process, \newtxt{and \refappendixCoalesce{sec:appdx} contains some example levels}.

Now, if we are interested in using \mcc with automatic level generators, they must satisfy some requirements. Specifically, the generators must generate levels quickly (as we need to generate many low-level structures); be able to generate structures of arbitrary sizes (as the sizes of the final level and its components are not fixed beforehand); and generate many diverse levels (to prevent the different structures from looking identical, and thus uninteresting, to a player).

Although our approach is agnostic to the exact level generators used, one method that is particularly well-suited to act as the low-level generator is \pcgnn, which satisfies all of our requirements. Additional benefits include that the training time is generally low~\citep{beukman2022Procedural} and, while we can include detailed domain knowledge, this is not required.

\begin{algorithm}
    \caption{\mcc}\label{alg:mcc}
    \begin{algorithmic}[1]
    \State
        \Comment{We call this initially with the top-level generator, as well as the desired size of the final level.}
        \Procedure{Generate}{$g \in G, \text{size} \in \mathbb{Z}^2 \cup \mathbb{Z}^3$}
          \State \Comment{Generate a map, possibly abstract, using $g$}
          \State map = \Call{MakeMap}{g, $size / S_g$}
          \If {$M_g$ is null}
                \State return map \Comment{Return map if this is the lowest level}
          \EndIf
          \State \Comment{Otherwise recurse to the next level in the hierarchy}
          \State \Comment{Create an empty map of sufficient size}
          \State FilledOutMap = Empty(size)
          \State \Comment{For each position, tile and coalesced size in the map}
          \ForAll {$pos, tile, {S}_{c} \in \Call{Coalesce}{\text{map}}$} 
            \State   \Comment{Recursively generate using the appropriate size}
            \State $\text{NewMap} = \Call{Generate}{M_g(tile), \text{size}=S_g \times {S}_{c}}$
            \State   \Comment{Place this generated map at the correct location}
            \State $\text{FilledOutMap}[pos \times S_g:(pos + {S_{c}}) \times S_g] = \text{NewMap}$
          \EndFor
          \State \Return FilledOutMap
        \EndProcedure
    \end{algorithmic}
\end{algorithm}

\subsection{Using \pcgnn in \mcc}\label{sec:method:pcgnn_plus_mcc}
Having presented the workings of \mcc, we next describe a concrete approach that uses \pcgnn to train the low-level generators.
First we train each component separately, using some objective function that describes a feasible individual. Once each of these generators has been trained, we can compose them together, needing only to specify which tiles map to which generators. This has multiple benefits; for instance, we can train and design multiple generators in parallel, obtaining a repertoire of components that can be composed later on. The approach is also modular, so we could leverage the same generator for different purposes (e.g., a house could be used in many different scenarios). 

Additionally, we can also train generators faster in this factored way compared to training one monolithic generator. 
Consider building an $n \times n$ level that consists of some high-level structure (e.g., an abstract map consisting of single-tile houses and roads) and low-level components (e.g., a generated house).
Let us compose the $n\times n$ level out of a high-level structure of size $\sqrt{n} \times \sqrt{n}$ and low-level components of size $\sqrt{n} \times \sqrt{n}$.
Now, during training, the flat method must generate levels of size $n \times n$. For the composed method, however, we have two separate training procedures, each generating levels of size $\sqrt{n} \times \sqrt{n}$. 
Thus, the overall time complexity\footnote{We omit factors such as the population size and the number of generations as they are kept constant for both approaches.} of training the flat method would be $O((n \times n)) = O(n^2)$, whereas the composed method would be $O(2(\sqrt{n} \times \sqrt{n})) = O(n)$.
For large levels, this speedup in training time can be quite significant. Note that during inference, however, each method generates the same amount of tiles.

Furthermore, as there is no dependence between different level generators during training, we can redefine the meaning of a specific generator. For instance, a town generates a tilemap containing houses, roads and gardens. We could also change the interpretation of the individual tiles, resulting in the same generator generating cities that consist of towns, highways and parks. This effectively constructs multiple different tree structures, with the same generator as the root and different child generators. In general, this approach would work as long as the same feasibility criteria are valid in both cases.

Finally, while \pcgnn is particularly suited to be used as the low-level generator in \mcc, this is by no means a requirement. We could use other suitable level generators (e.g., \citep{earle2021Illuminating,shu2021Experiencedriven,earle2021Learning}), hand-designed components, or a combination of these. This provides a significant amount of control and flexibility, which is useful in designing a game~\citep{lai2020Towards}.

\section{Experiments and Results}
\label{chap:experiments}
\subsection{Experimental Setup}
\subsubsection{Domains}
We consider two domains to demonstrate the capabilities of \mcc. We first quantitatively demonstrate the effects of composition in a simple 2D town-building game. We next experiment on \minecraft to demonstrate that our approach can also be applied to more complex 3D games.
\subsubsection{Experiments}
We perform three main experiments in this section. In \autoref{sec:exp:hierarchical}, we first illustrate the benefits of our hierarchical generation approach as the complexity of the lower-level structures increases. The second experiment, in \autoref{sec:exp:comp_vs_flat}, directly compares our hierarchical method against a non-compositional, flat approach. Finally, in \autoref{sec:showcase}, we qualitatively evaluate our method by generating complex towns and cities in \minecraft. \newtxt{More details about our \pcgnn implementation and our experiments, including hyperparameters and exact fitness functions, are listed in \refappendixExtsAndHyperparams{sec:appdx}, respectively.}
In our quantitative experiments, we assess the extent to which each method fulfils the fitness functions used during optimisation. This enables us to measure how leveraging hierarchy can streamline the optimisation process, resulting in improved fitness. Achieving a higher fitness value corresponds to the generator more faithfully realising the designer's intended outcome.

\subsection{Hierarchical Generation}\label{sec:exp:hierarchical}

A compositional approach allows us to simplify the task of the top-level generator by abstracting away the details of the lower-level components. In particular, for a compositional approach to build a house, it requires a single action---placing the high-level house tile. A non-compositional approach, however, requires a large number of coordinated actions---placing each of the low-level tiles that make up the house.
In this experiment we compare these two paradigms, using a simple town level with houses, gardens and roads.

To this end, we introduce the concept of \textit{window size}, a proxy for how complex the lower-level structures are. For instance, a window size of $1\times 1$ means that placing a single tile is sufficient to generate a high-level structure, corresponding to the compositional setting. A larger window size, such as $2\times 2$, means that a house requires $4$ coordinated actions to build, which corresponds to a non-compositional approach. Similarly, $5\times 5$ means that a house requires $25$ coordinated actions, i.e., the low-level structures are harder to build.

\begin{figure}[h]
  \centering
  \includegraphics[width=0.95\linewidth]{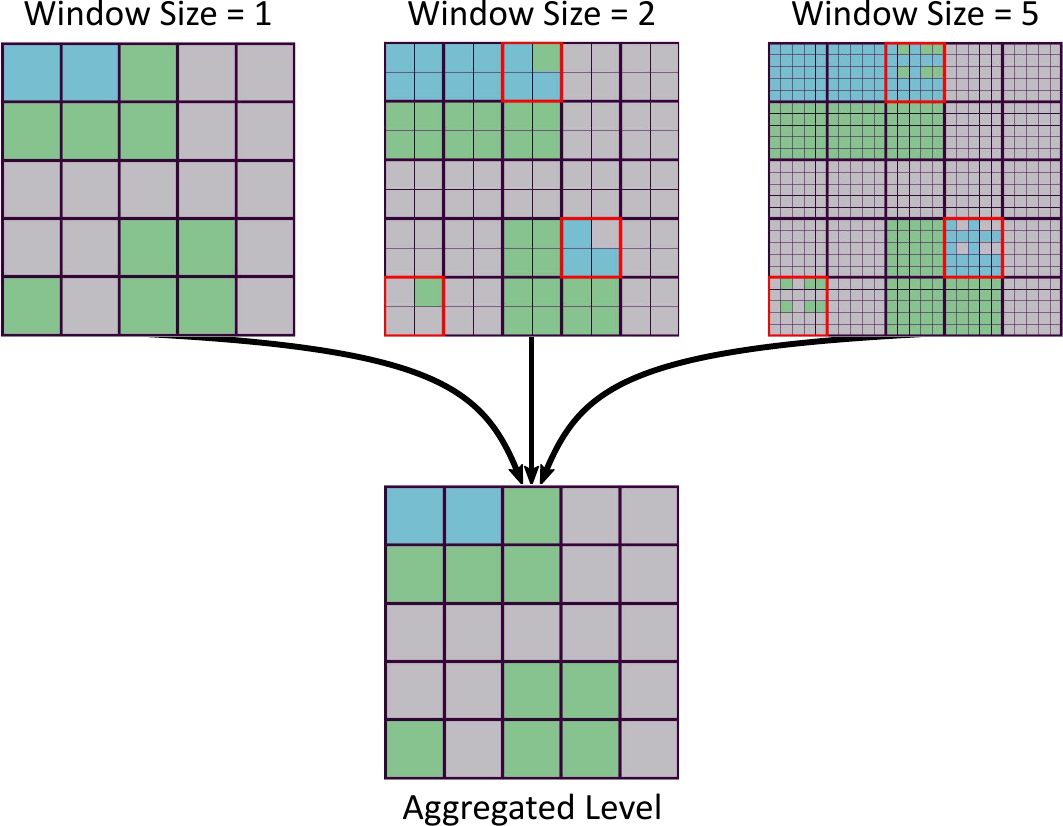}
  \caption{The top row illustrates three generated levels, with window sizes of $1\times 1, 2 \times 2$ and $5\times 5$, respectively. The bottom row shows the aggregated level, i.e., the level that each of the above ones reduces to. The red squares indicate logical tiles that were collapsed to the default tile of grass due to not being fully populated with the same tile. Here, blue tiles are houses, green tiles are gardens and grey corresponds to roads.}\label{fig:exps:windowsize:all}
\end{figure}

We implement this as follows (see \autoref{fig:exps:windowsize:all} for an example): For a window size of $2\times 2$, say, we consider each non-overlapping $2 \times 2$ window as a single tile. If all the tiles within a window of a certain size are identical, then a single tile of that type is placed in the aggregated level. Otherwise, the tile is replaced with a ``default'' tile---gardens in this case. While we could change this rule (e.g., taking the majority tile), our choice simulates that a certain number of coordinated actions \textit{must} be performed to generate a single component. By contrast, in the compositional case, only one action is required. In essence, the larger the window size, the more coordinated actions are required to generate a single tile. We also consider window sizes of $3\times 3$, $4\times 4$, $5\times 5$ and $10\times 10$, to evaluate how the complexity of the low-level structures influences our final performance. Note that we generate levels with an aggregated size of $10\times 10$ tiles. This means that, for a window size of $2\times 2$, the number of total tiles is $20\times 20$, which is then aggregated into a $10\times 10$ level (where each $2\times 2$ block in the large level is compressed to a single tile).

This experimental setup has the added benefit of being comparable, as each method uses the same fitness functions, and the only difference is in the size of the generated structures.
The fitness functions used here are (1) reachability, where houses must be reachable via roads, and (2) a probability fitness, incentivising a roughly equal number of house, garden and road tiles. See \refappendixHyperParams{sec:appdx} for more details.

\begin{figure}
  \centering
  \includegraphics[width=1\linewidth]{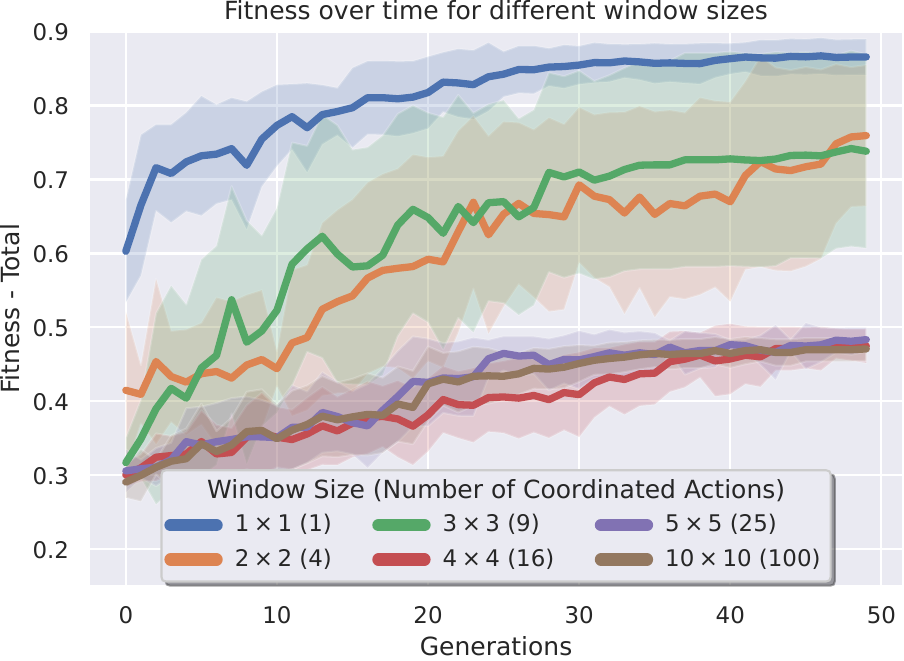}
  \caption{A plot showing the maximum fitness value in the population (i.e., how ``correct'' the generated levels are) over time for different window sizes (\newtxt{i.e., how many coordinated actions are required to build the lower-level structures}). The mean over 10 seeds is shown, with standard deviation shaded.}
  \label{fig:mcc:results:composition}
\end{figure}

We compare the fitness of each method over time in \autoref{fig:mcc:results:composition}, which indicates that using the compositional approach results in a higher final fitness value compared to the other, non-compositional methods. The performance decreases significantly when we have a much larger $10\times 10$ window size (corresponding to an overall level size of $100\times 100$), where each house requires $10^2 = 100$ coordinated actions to build. Even this is still significantly fewer actions than a real \minecraft house. Overall, the results show that simplifying the task of the top-level generator by abstracting away the details of the lower-level components is beneficial.

\subsection{Composition vs Flat Generation}\label{sec:exp:comp_vs_flat}
Our next experiment directly compares \mcc against a non-compositional baseline that generates the entire level using \pcgnn. In particular, we are again interested in building towns, this time attempting to replicate a specific, randomly generated, town layout.

For the compositional approach, we learn two separate generators for houses and towns, while the non-compositional method generates the entire level at once. In particular, we have a $25\times 25$ level, consisting of a collection of $5\times 5$ houses. The high-level town structure also has a size of $5\times 5$. The fitness functions used did not include novelty, and measured the average overlap between the desired level and the generated one. The desired levels use one set layout for houses, where walls surround empty space, and a randomly generated town layout. We generate 20 of these random town layouts, train a separate generator on each of them, and average the fitness values obtained. 

\begin{figure}
  \centering
  \includegraphics[width=1\linewidth]{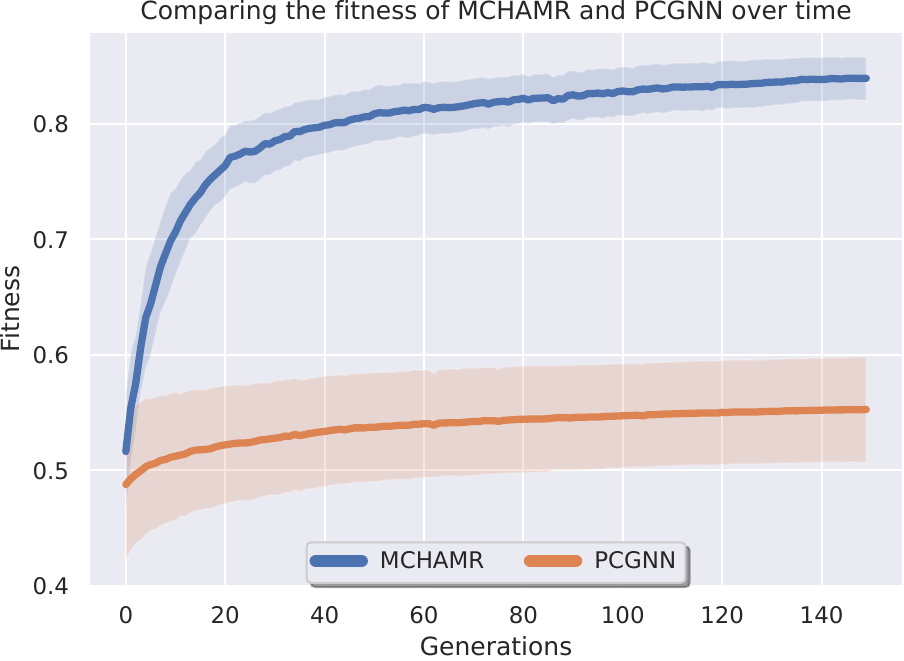}
  \caption{Illustrating the fitness over time on 20 random town layouts for \mcc (consisting of the composed town and house generators) compared to the flat, non-compositional method, corresponding to vanilla \pcgnn. 
  We first compute the average fitness across 10 seeds and plot the mean and standard deviation over the 20 layouts.}
  \label{fig:mcc:results:composed_vs_not}
\end{figure}

The fitness results over time are shown in \autoref{fig:mcc:results:composed_vs_not}, where it is clear that \pcgnn's fitness cannot touch that of \mcc. Our composed method achieves significantly higher fitness, and generates much more accurate levels than the flat method. See \refappendixHyperParams{sec:appendix} for examples of the generated levels.

\newcommand{\mikefigures}{\begin{figure*}[h!]
  \centering
  \begin{minipage}[t]{0.49\linewidth}
      \includegraphics[width=1\linewidth]{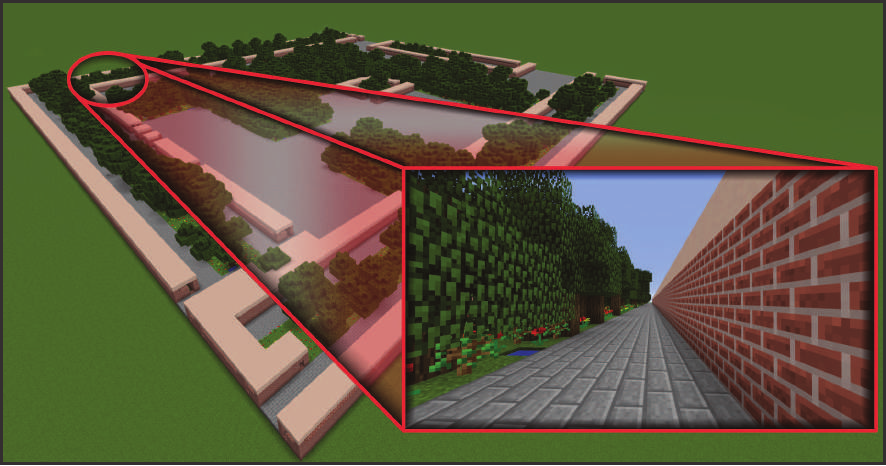}
    \caption{A generated city and a zoomed-in portion of an alleyway. The city and town generators are the same.}
    \label{fig:mcc:showcase:subfigs}
  \end{minipage}\hfill
  \begin{minipage}[t]{0.49\linewidth}
    \includegraphics[width=1\linewidth]{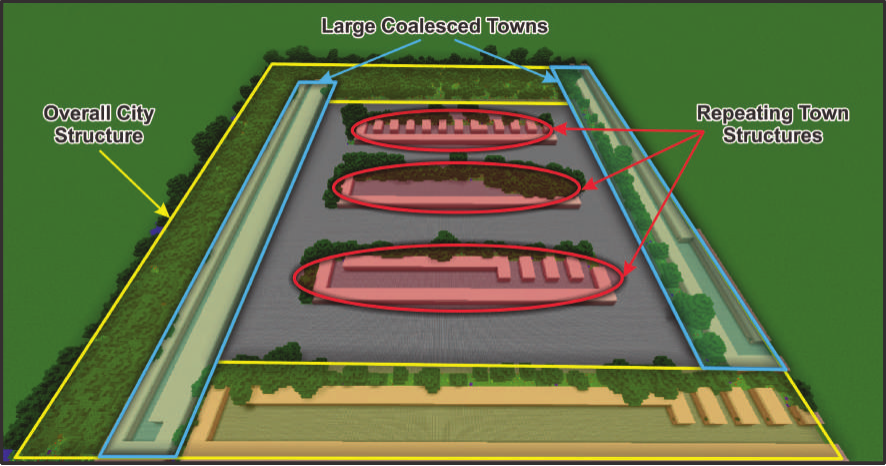}
  \caption{An annotated example of a city.}
  \label{fig:mcc:showcase:annotate}
\end{minipage}
\end{figure*}
\newcommand{\mikehfill}{ \hspace*{0.0001in}}
\newcommand{\wwww}{0.16}

\renewcommand{\wwww}{0.49}
\begin{figure*}[h!]
  \centering
  
  \begin{subfigure}[t]{\wwww\linewidth}
    \includegraphics[width=1\linewidth]{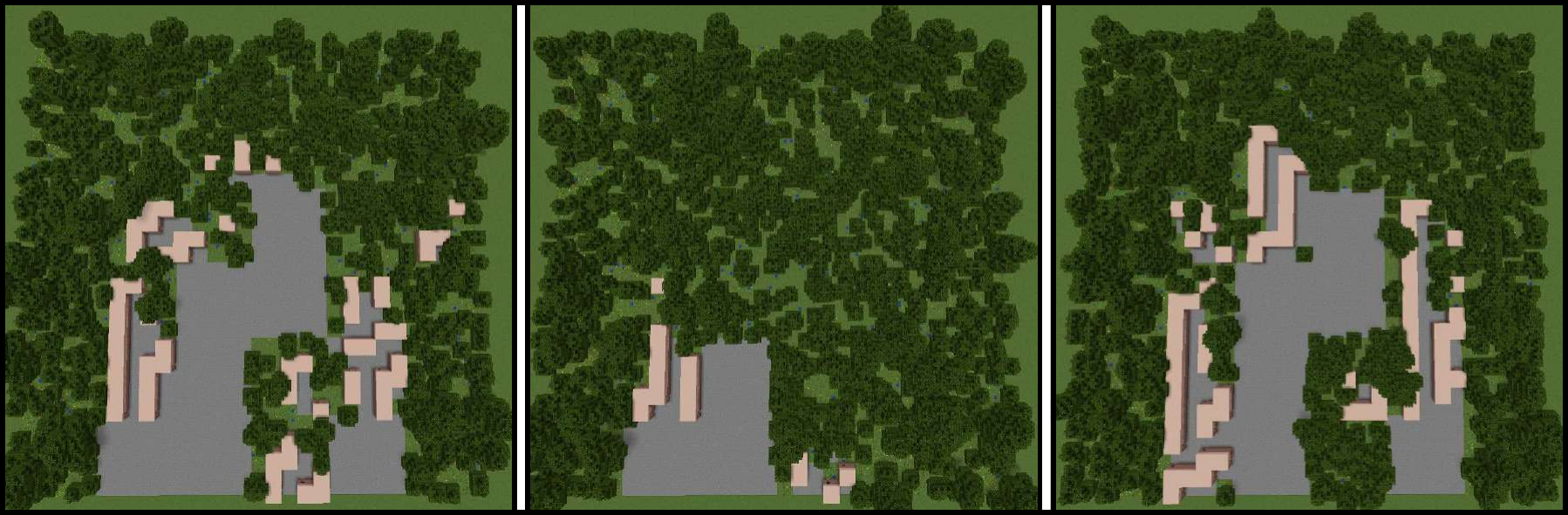}
    \caption{Three levels from generator A.}
  \end{subfigure}\hfill
  \begin{subfigure}[t]{\wwww\linewidth}
    \includegraphics[width=1\linewidth]{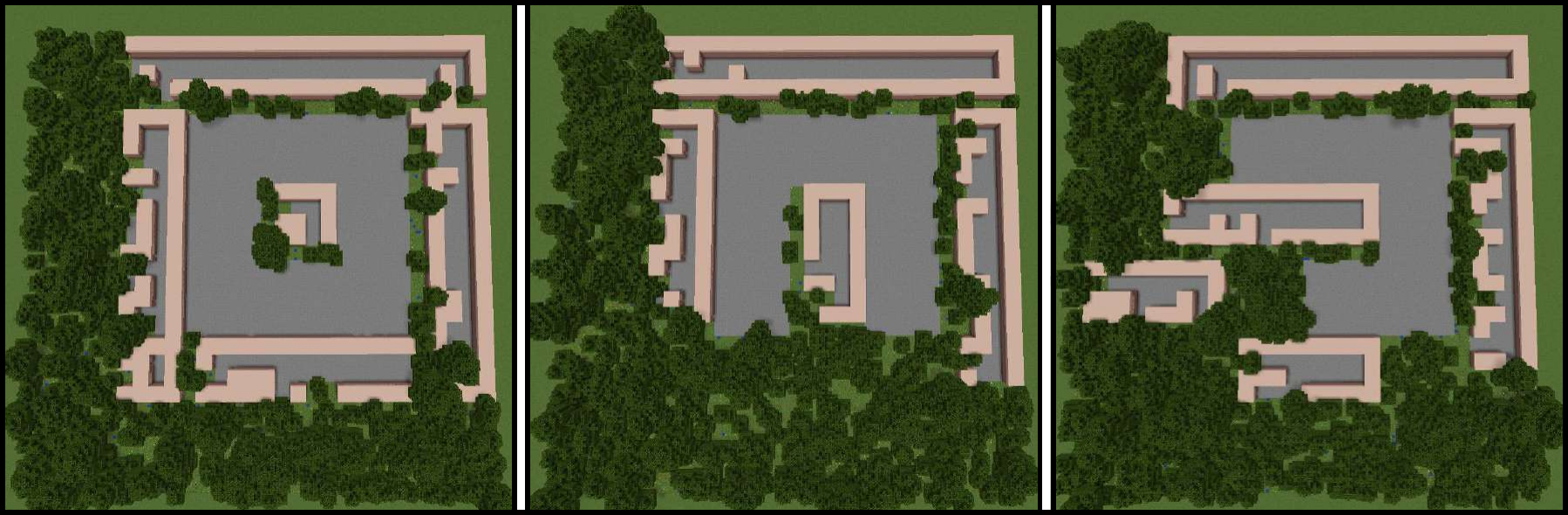}
    \caption{Three levels from generator B.}
  \end{subfigure}
  \caption{Showing three levels each from two generators (a) and (b). Each generator comes from the same experiment, just with different initial random seeds. Generator A has large swaths of gardens, with some towns around the center. Generator B generally has gardens in the bottom and left sides, with towns in the middle and along the top and right edges.}
  \label{fig:mcc:showcase:diversity:all}
\end{figure*}}

\subsection{Minecraft Showcase}\label{sec:showcase}
\mikefigures
Finally, we evaluate our method on \minecraft and show a few qualitative examples of interesting and complex structures that were generated by \mcc. We use the Evocraft~\citep{grbic2020Evocraft} library to place and visualise the generated structures in game.\footnote{\url{https://github.com/real-itu/Evocraft-py}}

Here we generate 3D settlements, where each town consists of houses, gardens and roads. Additionally, to demonstrate the notion of reusing a single generator, we further generate \textit{cities} using the town generator, by redefining houses to be towns. Concretely, we use the following generators, with \refappendixHyperParams{sec:appendix} discussing the fitness functions in more depth.

\begin{LaTeXdescription}
  \item[Town] A high level 2D tilemap of a town. \newtxt{We used several objectives, but (similarly to \autoref{sec:exp:hierarchical}) primarily rewarded reachability and having roughly an equal number of house, road and garden tiles.}
  \item[House] A 3D house, \newtxt{incentivised to have a roof, walls and an empty interior.} The house also had novelty and intra-generator novelty objectives.
  \item[Garden] A 2D tilemap with flowers, grass, ponds and trees. \newtxt{The generator is incentivised to generate levels with at least one of every tile, no more than 5\% pond tiles, between 20\% and 70\% grass tiles and trees that are not too close to each other.} Gardens and houses share the same novelty objectives.
\end{LaTeXdescription}

In \autoref{fig:mcc:showcase:subfigs}, we use the same generator for the city and the town. We also show a first-person view of a specific section to illustrate the detail in each part of the generated settlement.

\autoref{fig:mcc:showcase:annotate} shows an annotated example of a generated city. The large, high-level structure of the city is generated by the top-level generator. Then, each tile that it places is transformed into a town by the town generator. We see that coalescing results in longer towns and the towns themselves, while not identical, contain similar repeating structures.

In \autoref{fig:mcc:showcase:diversity:all}, we illustrate the benefits of incentivising novelty during training. In this figure, we use the same houses and garden generators as before, alongside a town generator that was trained using four fitness functions: (1) Intra-generator Novelty; (2) Novelty Search; (3) Reachability; and (4) a Probability fitness incentivising roughly an equal number of houses, gardens and roads. We use the tile-pattern KL-Divergence metric, with a pattern size of $2\times 2$, as the novelty distance function~\citep{lucas2019Tile} and weigh the fitness functions with ratios $1:1:4:4$.
We generate cities using the same neural network as the towns.
These levels show that if we use the same generator to generate multiple levels, we can obtain qualitatively different structures while still adhering to a certain style. Furthermore, running the same training procedure multiple times---with the same hyperparameters but a different initial random seed---can result in very different structures.
Details about training, novelty objectives and a comparison of novelty scores can be found in \refappendixQuantNov{sec:appendix}.

Finally, we note that the structures we generate are very large in the scale of standard PCG. For instance, \textit{Super Mario Bros.} levels have size $114 \times 14$. Much of the work in \minecraft also generates relatively small structures, such as $7\times 7\times 7$ mazes~\citep{jiang2022Learning} or $20\times 20 \times 20$ lattices~\citep{barthet2022openended}. The composed towns we generate can easily be $200\times 5\times 200$ or larger, totalling more than $100\text{ }000$ blocks. Despite this massive scale, we are still able to generate coherent structures and sensible towns. Large settlements are also generated for the GDMC competition~\citep{salge2018Generative}, but unlike our approach, these generations often require significant hand-designing effort~\citep{salge2021Settlement}.

\section{Discussion}
The approach introduced here, \mcc, follows a recent trend that emphasises the importance of more complex and open-ended creations in PCG~\citep{salge2018Generative,grbic2020Evocraft,barthet2022openended}.
We break the generation problem down into separate components, and train a model for each component. This is different from much of the contemporary research on generative models, where methods tend to train one monolithic model. We believe that there are several good reasons for this factorisation. First, by factorising each salient component of a level, each of these can be changed separately without altering anything else. 
For instance, suppose we have a good town generator that we are satisfied with, but wish to change the house layout. If we had one large model, it would need to be entirely retrained with an alternative objective function, with no guarantee that the other pieces would indeed remain unchanged. On the other hand, when we factorise the components, each component can be changed independently. 
Secondly, if we have separate generators, each generator's fitness function is much simpler to create (for the designer) and also simpler to optimise (for the learning algorithm). By contrast, it can be challenging to design a single monolithic objective function for complex tasks such as building a town. Even if we can create such an objective, it would likely be hard to optimise by the agent.

Our approach does require a designer to specify the components and desired hierarchy, as well as particular fitness functions. We believe that this gives the designer immense control over the generation, as they can completely specify which components should be used, and how they should be combined. Additionally, we believe that designing a fitness function for each component---while requiring some effort---is an elegant way to allow designers to specify what is defined as good, without needing to design the components themselves.

Our method has numerous applications, such as (partially) generating large open-world games. This would allow designers freedom in specifying certain, reusable low-level components that can be composed in numerous different ways, thereby enticing players towards exploration.

Furthermore, in contrast to many other methods, we abstract away details relevant to a particular game by having separate high- and low-level generators. Separating these responsibilities could enable the abstract generators to be used in a variety of games, needing only to swap out the low-level generators for each game.
Relatedly, since we independently train the generators, it would be possible to have a community-driven database of useful generators, which can then be seamlessly composed together to generate novel artifacts. This could be similar to \textit{Picbreeder}~\citep{secretan2011Picbreeder},\footnote{\url{https://nbenko1.github.io/}} where users collectively contribute to the generation of many interesting images.

\section{Limitations and Future Work}

Our method is just a first step in this direction, and there are numerous avenues for future work. One promising avenue would be to use a combination of different PCG methods as low-level generators~\citep{khalifa2020Pcgrl,earle2021Illuminating,jiang2022Learning}, instead of just \pcgnn.
For instance, we could use data-based approaches to generate the low-level components (using, e.g., existing house datasets), while generating the high-level structure using a fitness-based method.
Furthermore, while our method is designed particularly with tile-based games in mind, we believe that it could be extended to other types of games by leveraging more appropriate low-level generators.
Our tile-based focus may additionally lead to discretisation artifacts, as components can only be placed on a rectangular grid. While this simplifies the hierarchical computation, it may lead to blocky-looking structures. Coalescing is one way to partially alleviate this problem: by generating at a higher resolution and coalescing similar tiles, we could obtain smoother shapes. 
Future work could also explore other techniques to address this limitation.

While the coalescing rules we used were intentionally simple, they can be made arbitrarily complex; for instance, a designer may wish to only coalesce under certain conditions, and leave most connected tiles as single components.

In future work, we would like to explore the use of quality-diversity algorithms~\citep{mouret2015Illuminating} to obtain a collection of diverse and high-performing low-level generators, and use these interchangeably to obtain more diverse content. Furthermore, since we use a 2D generator as the root, the generated levels do not have any verticality. Adding vertical aspects to the levels (e.g., mountains, multi-story houses, etc.) would thus be valuable.

\newrev{Finally, we have a clear separation between each component without any explicit communication between components, be it at the same or across different hierarchical levels. Thus, all structure must come from the top-level generator, similar to the top-down approach outlined by \citet{liapis2019Orchestrating}, and no emergent structure is generated via the interaction of the low-level generators. 
This may carry some disadvantages, such as neighbouring components being oblivious of one another, leading to incoherent generations.

However, in our approach, this problem could be circumvented by adding more low-level components, and expanding the tileset of the top-level generator---thereby increasing its complexity. The tradeoff here is that this may require more human effort in designing these additional generators (via their fitness functions), and altering the existing hierarchy.

Despite these limitations, this independence enables us to learn modular components that can be used in several settings, whereas reuse would be more difficult if there is a tight coupling between components. 
Furthermore, we allow the designer to specify the structure and the high-level generator in charge of implementing it. This gives the designer more control compared to a case where the only structure is obtained by the unpredictable interaction of the low-level generators.
}

Relatedly, independently training the generators does not enable them to learn how to best combine their generated components, or for one generator to make up for the shortfall of another. 
Instead, the designer specifies the hierarchy and fitness functions for each component. This shifts some of this responsibility onto the designer, but avoids the problem where the high- and low-level generators are incompatible. Future work could consider modifications to this, where generators are jointly trained.
This idea could also relate to more open-ended avenues, where the generated structures increase in complexity over time, as opposed to reaching an endpoint when the feasibility fitness is maximised~\citep{grbic2020Evocraft,barthet2022openended}. One way to achieve this could be learning how to dynamically compose different generators, or using learned objective functions.

\section{Conclusion}
We introduce \mcc, a compositional approach to level generation, which leverages multiple simple generators to generate large and complex structures. Our approach has multiple benefits, such as (1) being configurable, allowing the designer to choose which low-level generators are used; (2) it being straightforward to combine different generators, without needing to specify one monolithic fitness function; and (3) simplifying each generator's task by decomposing the problem, potentially allowing for less training or smaller (and thus faster) models. We demonstrate that our approach improves generation compared to a non-compositional approach, and that it is able to generate large-scale and complex settlements in \minecraft. 
Ultimately, we hope that this work is a step towards more complex and large-scale generations, which is necessary for the adoption of PCG in the gaming industry.

\section*{Acknowledgments}
\noindent Computations were performed using HPC infrastructure provided by the MSS unit at the University of the Witwatersrand. We thank the reviewers for their insightful comments, which helped to strengthen the final version of this paper.

\bibliographystyle{IEEEtran}
{
\footnotesize
\bibliography{IEEEabrv.bib,clean.bib}
}
\appendices
\label{sec:appdx}
\section*{Appendix}
We structure the supplementary material as follows: \refappendixExt{sec:appdx:extensions} discusses the extensions and improvements we made to \pcgnn. \refappendixRepr{sec:appdx:repr} details some reproducibility details and our experimental setup. Next, \refappendixHyperParams{sec:appdx:hyperparams} lists the hyperparamters and in-depth definitions of the fitness functions we used for each experiment. It also shows some examples of the generated levels for our quantitative experiments. \refappendixCoalesce{sec:appdx:coalesce} contains example levels in \minecraft with and without coalescing. Finally, \refappendixQuantNov{sec:appdx:novresults} contains more details about the qualitative novelty results in the main text. In this section, we also quantitatively compare the effect of several different novelty distance functions and show some example levels when training with each distance function.
\section{Extensions to \pcgnn}
\label{sec:appdx:extensions}
We extend standard \pcgnn by adding in additional features useful for \mcc. The specific features we use are as follows:
\begin{LaTeXdescription}
    \item[3D] We simply generalise the generation procedure to three dimensions, by considering a level as a $W\times H\times D$ block of tiles, and iterating through all of the tiles.
    \item[Multiple Iterations] Using multiple iterations is also a relatively simple change: after performing the generation once, we use this as a starting point, and repeat the process for multiple iterations. The intuition here is that performing this process multiple times has the effect of increasing the effective \textit{receptive field} of the network. This is similar to how deep convolutional neural networks are able to focus on long-range dependencies by stacking local interactions~\citep{luo2017Understanding}. An approach like this has been used successfully in PCG before; specifically, \citet{earle2021Illuminating} perform multiple processing iterations of a neural cellular automaton to generate levels.
    \item[Input Center Tile] This adds in the center tile the model is currently focusing on as an input. This is beneficial in the case of multiple iterations, so that the network can pass along information to the subsequent iterations.

    \item[Various Starting Points] While we mostly still use random initialisations, some experiments instead initialised the level with a specific ``default'' tile, so that the starting points are all the same.
\end{LaTeXdescription}
\section{Reproducibility Details and Experimental Setup}
\label{sec:appdx:repr}
As mentioned in the main text, we publicly release our source code at \url{\ghurl}. Implementation details, as well as instructions to reproduce our results, can be found there.

When plotting the fitness value in our experiments, we use the maximum fitness in the current population. This is done because we use the best individual at the end of training as our chosen generator. Furthermore, all of our fitness functions are scaled between 0 and 1, so $1.0$ corresponds to the optimal fitness.
\section{Hyperparameters}
\label{sec:appdx:hyperparams}
Here we detail the hyperparameters and exact setups used to generate the images in the main paper. If a particular parameter is not mentioned, the default value from \citet{beukman2022Procedural} was used; for example, all of our experiments padded the boundaries with $-1$. 
\subsection*{Different sizes of hierarchy}
\label{sec:hierarchy_vals}

\begin{table}[h]
    \centering
    \caption{Hyperparameters where we experimented with different window sizes, i.e. differing complexity levels of the base structures.}
    \label{tab:appdx:hierarchy}
\begin{adjustbox}{width=1\linewidth}
    \begin{tabular}{ll}
    \toprule
    Name & Value \\
    \midrule
    One Hot Encoding & True \\
    Generations & 50 \\
    Population Size & 50 \\
    Number of Random Variables & 1 \\
    Random Perturb Size & 0 \\
    Number of Iterations & 3 \\
    Input Center Tile & Yes \\
    Default Tile & Garden \\
    Novelty & No Novelty \\
    Level Starting Point & Default Tile (Garden) \\
    Number of levels for fitness calculations & 5 \\
    \bottomrule
\end{tabular}
\end{adjustbox}
\end{table}

\begin{figure*}[h!]
    \centering
    \begin{subfigure}{0.49\linewidth}
        \includegraphics[width=1\linewidth]{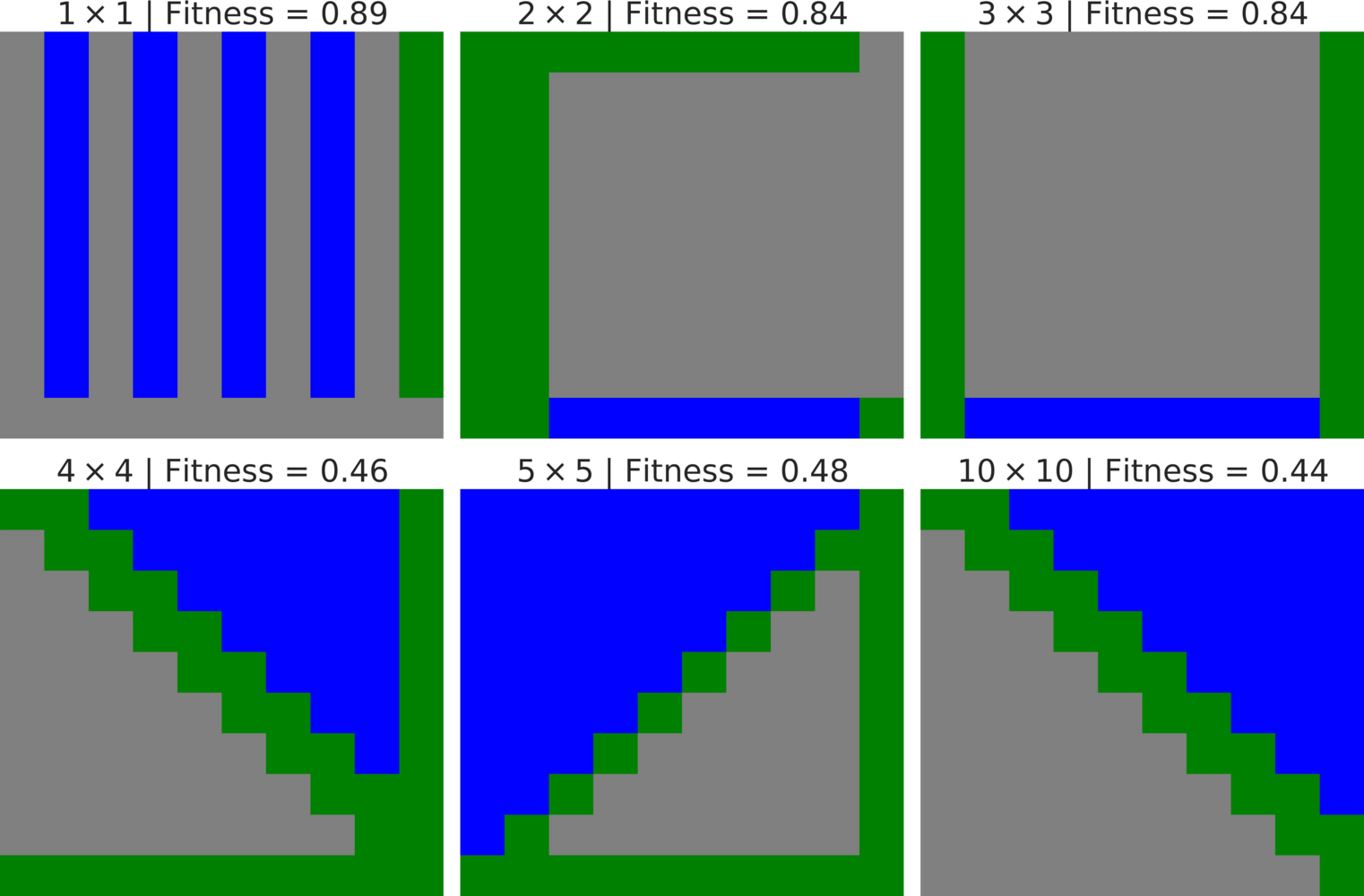}
        \caption{}
    \end{subfigure}
    \begin{subfigure}{0.49\linewidth}
        \includegraphics[width=1\linewidth]{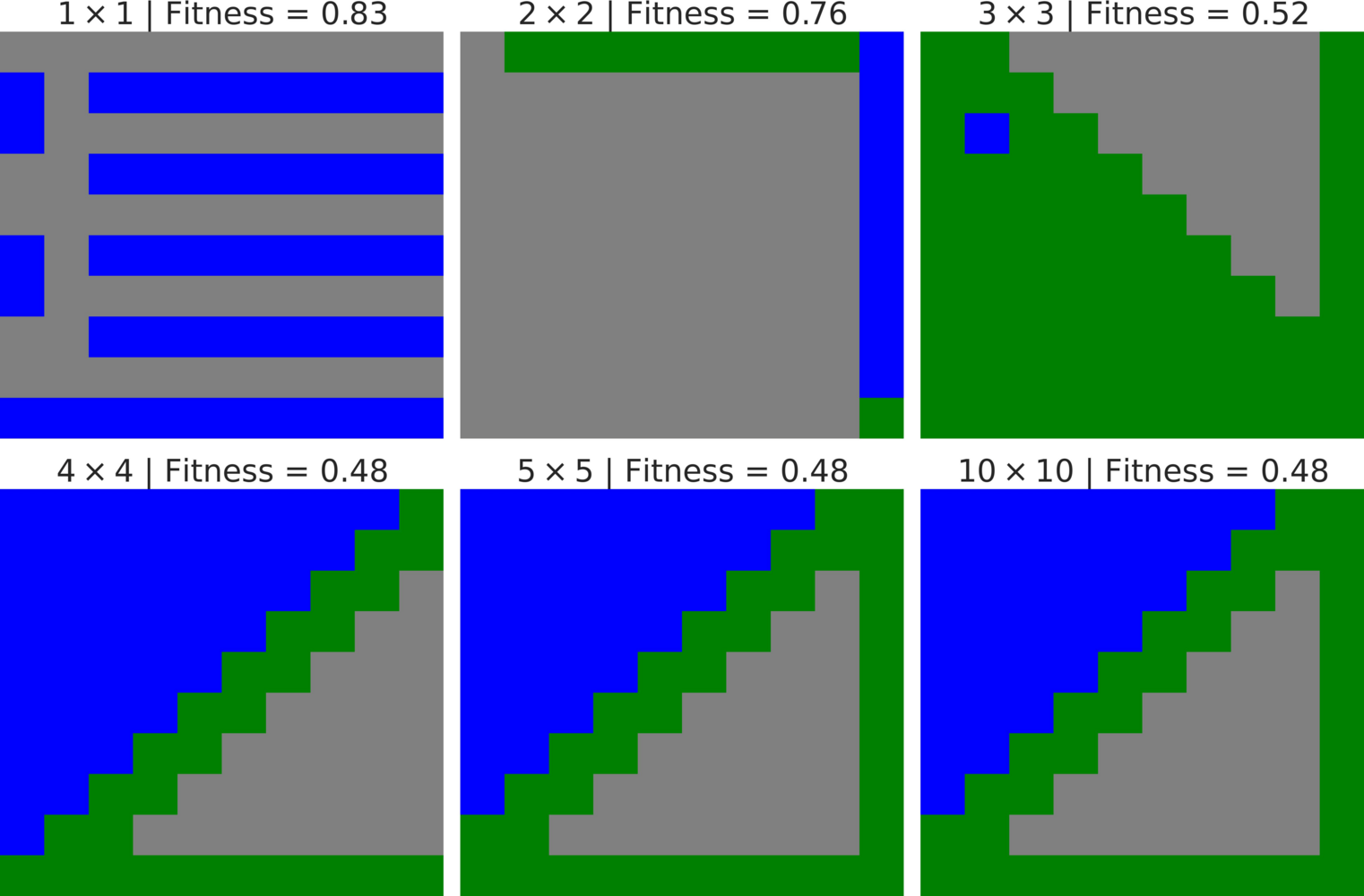}
        \caption{}
    \end{subfigure}

    \caption{Showcasing example levels generated by the networks with different logical tile sizes for two different seeds.}
    \label{fig:appx:tile_sizes_examples}
\end{figure*}

In this experiment, we compared the results when using different logical window sizes, from $1\times 1$, representing a composed approach, to $10\times 10$. 
For all of these experiments, we used the hyperparameters shown in \autoref{tab:appdx:hierarchy}, with \autoref{fig:appx:tile_sizes_examples} containing some examples of generated levels. We calculated the individual's fitness as the average of the following two fitness functions:
\label{sec:reach_fit}
\begin{LaTeXdescription}
    \item[Probability]: We compute the level's tile distribution, representing the frequency of each tile type in the level. This fitness is the square root of the Jensen-Shannon divergence~\citep{fuglede2004JensenShannon} between the level's tile distribution and an ``ideal'' distribution of 40\% house, 30\% garden and 30\% road.
    \item[Reachability]: All houses must be reachable (with allowable moves being up, down, left and right) from all other houses via roads without being directly connected. This fitness is calculated as $a \times b$, where $a = \min{(\frac{H}{20}, 1)}$, with $H$ being the number of houses that have between 1 and 3 (inclusive) neighbouring road tiles. $b = \frac{1}{(d_{house,road} + 1)(d_{road} + 1)}$, where $d_i = \min{(|1 - c_i|, 10)}$ and $c_i$ is the number of disconnected regions of tile type $i$. For example, if a level consists entirely of roads, with a vertical line of gardens through the middle of the level, then there would be two disconnected regions of road (to the left and right of the vertical line).
     When calculating $c_{house,road}$, we label houses and roads as the same tile, and find the number of disconnected regions (i.e. those regions separated by gardens). All of these operations can be efficiently implemented using morphological image operations, notably image labelling.\footnote{\url{https://scikit-image.org/docs/stable/api/skimage.morphology.html\#label}} Here, $a$ incentivises each house to be reachable by a road, without being completely isolated from gardens or other houses; $b$ specifies that there must be one connected road region, and that each house must be reachable from every other house (as $b$ is maximised when $c_i = 1$, corresponding to there only being one connected region).
\end{LaTeXdescription}

\begin{figure*}[h!]
    \centering
    \begin{minipage}[t]{.47\textwidth}
        \centering
        \begin{subfigure}[t]{0.32\linewidth}
            \includegraphics[width=1\linewidth]{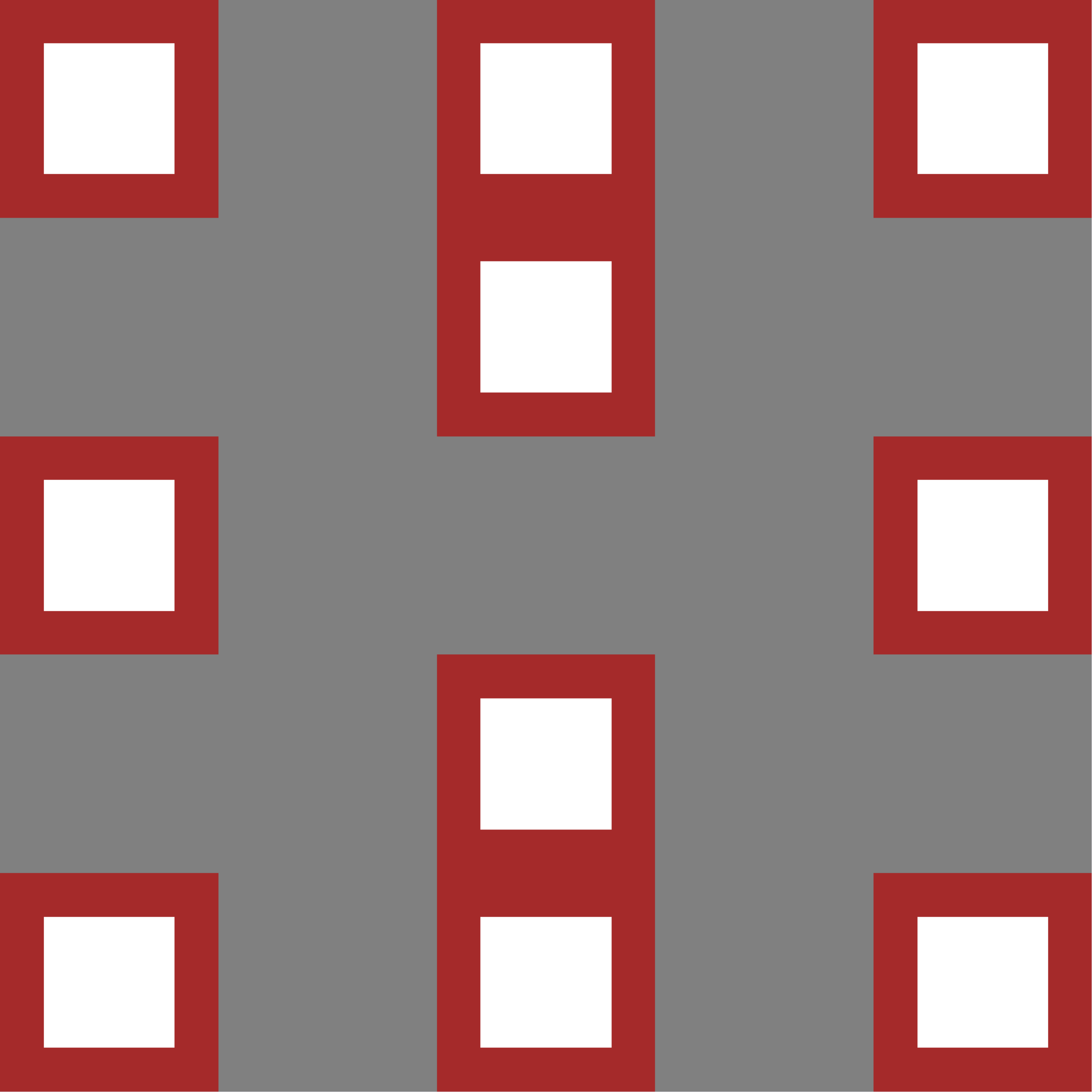}
            \caption{}
        \end{subfigure}
        \begin{subfigure}[t]{0.32\linewidth}
            \includegraphics[width=1\linewidth]{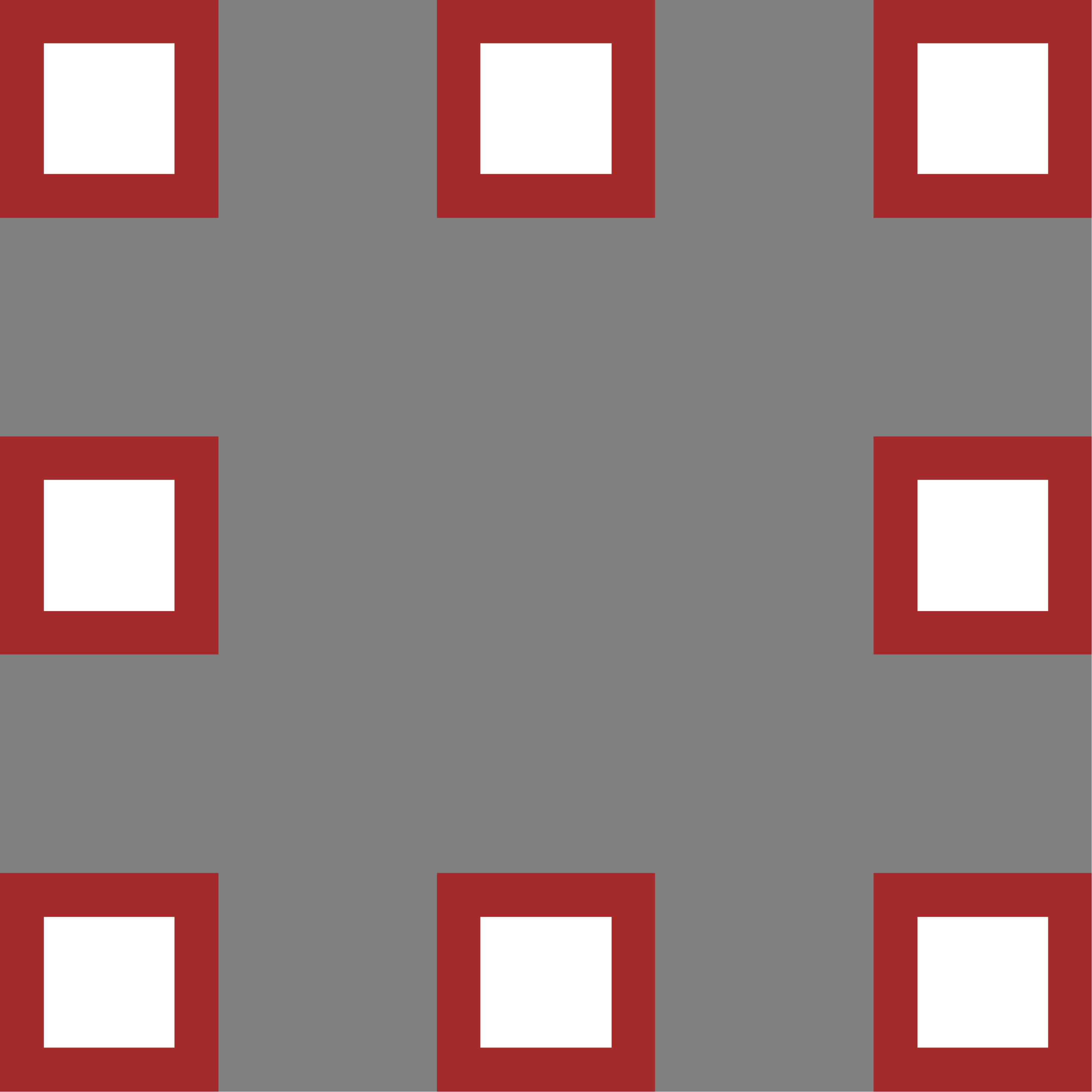}
            \caption{}
        \end{subfigure}
        \begin{subfigure}[t]{0.32\linewidth}
            \includegraphics[width=1\linewidth]{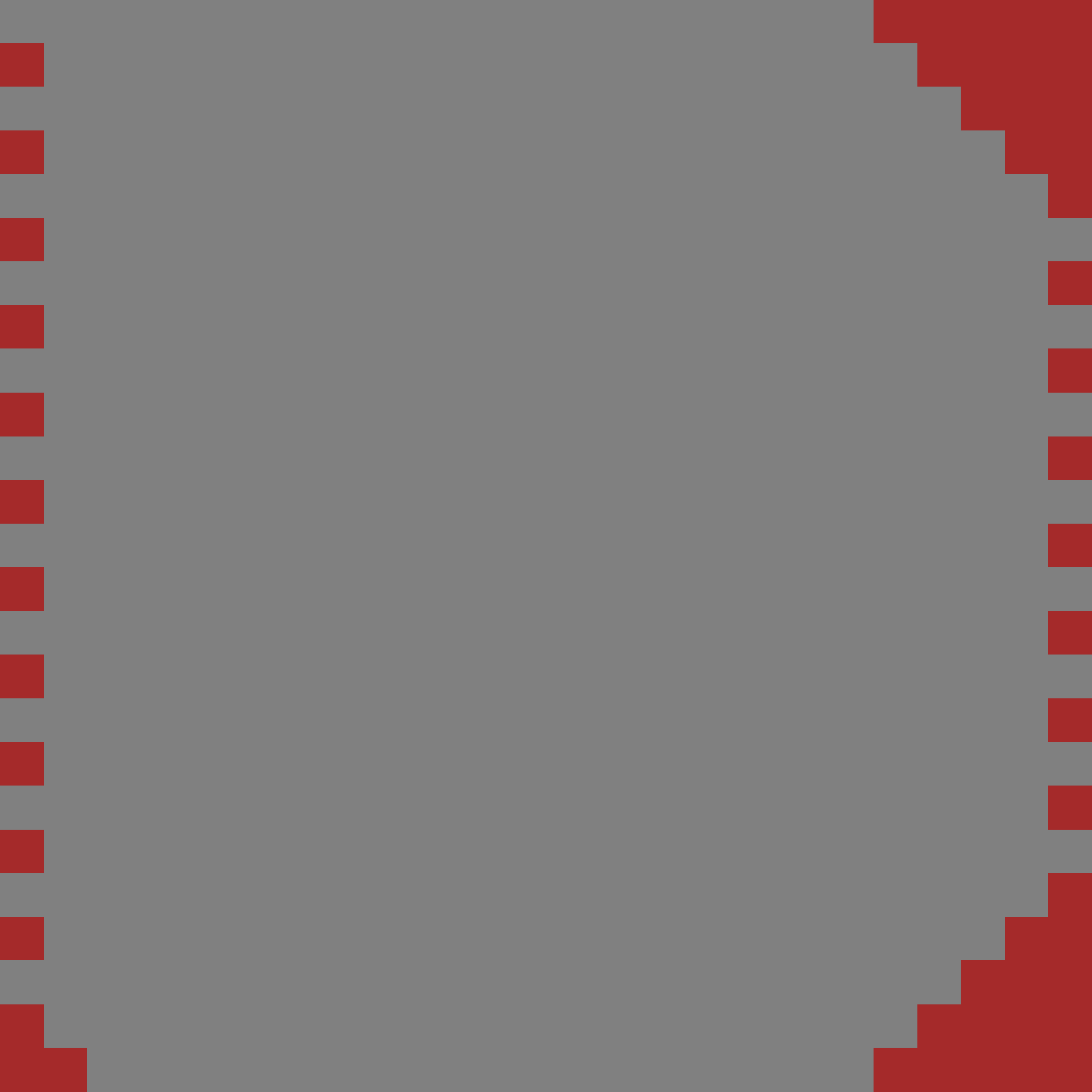}
            \caption{}
        \end{subfigure}
        \caption{}
        \label{fig:appx:compose_examples:hand}
    \end{minipage}
    \hfill
    \begin{minipage}[t]{.47\textwidth}
    \centering
    \begin{subfigure}[t]{0.32\linewidth}
        \includegraphics[width=1\linewidth]{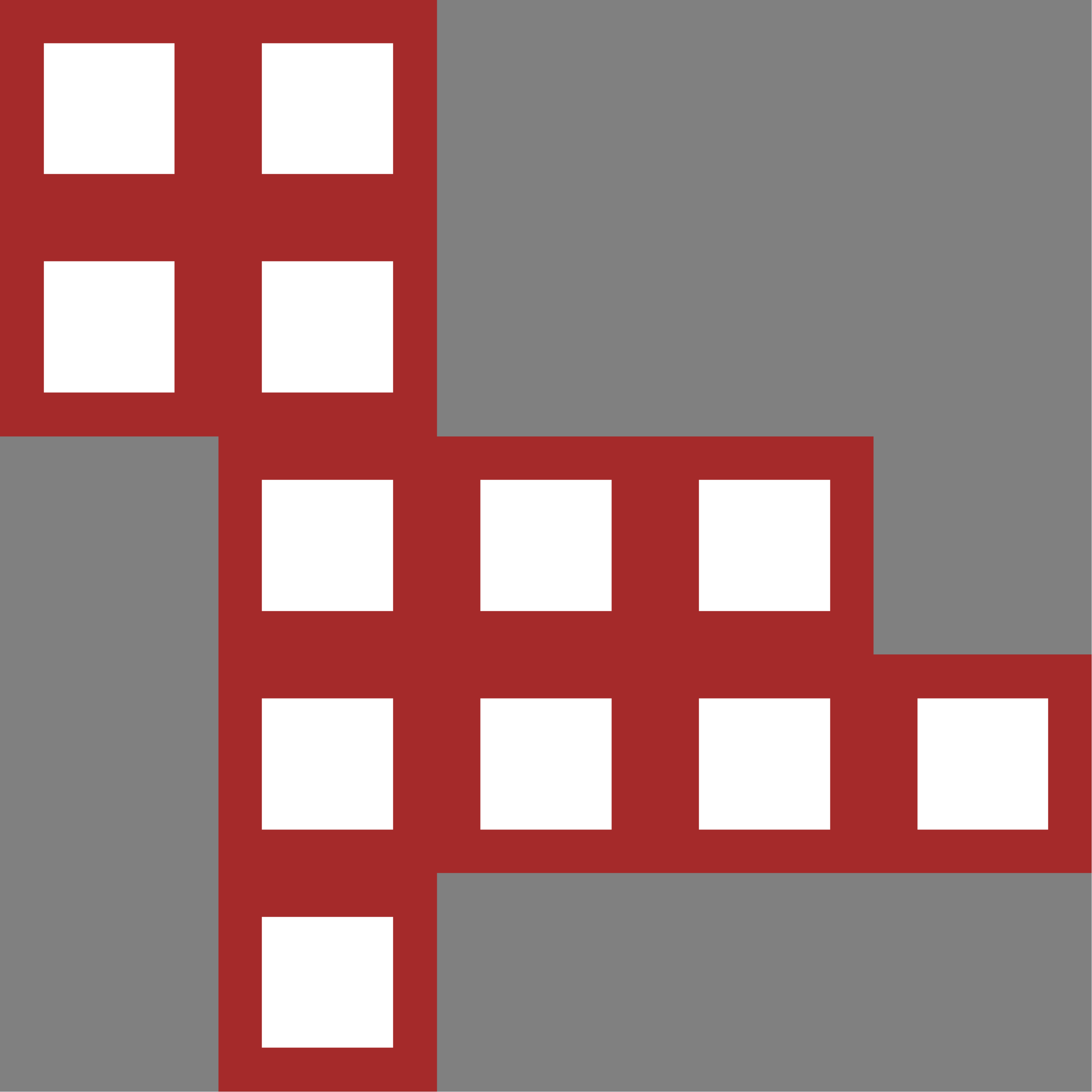}
        \caption{}
    \end{subfigure}
    \begin{subfigure}[t]{0.32\linewidth}
        \includegraphics[width=1\linewidth]{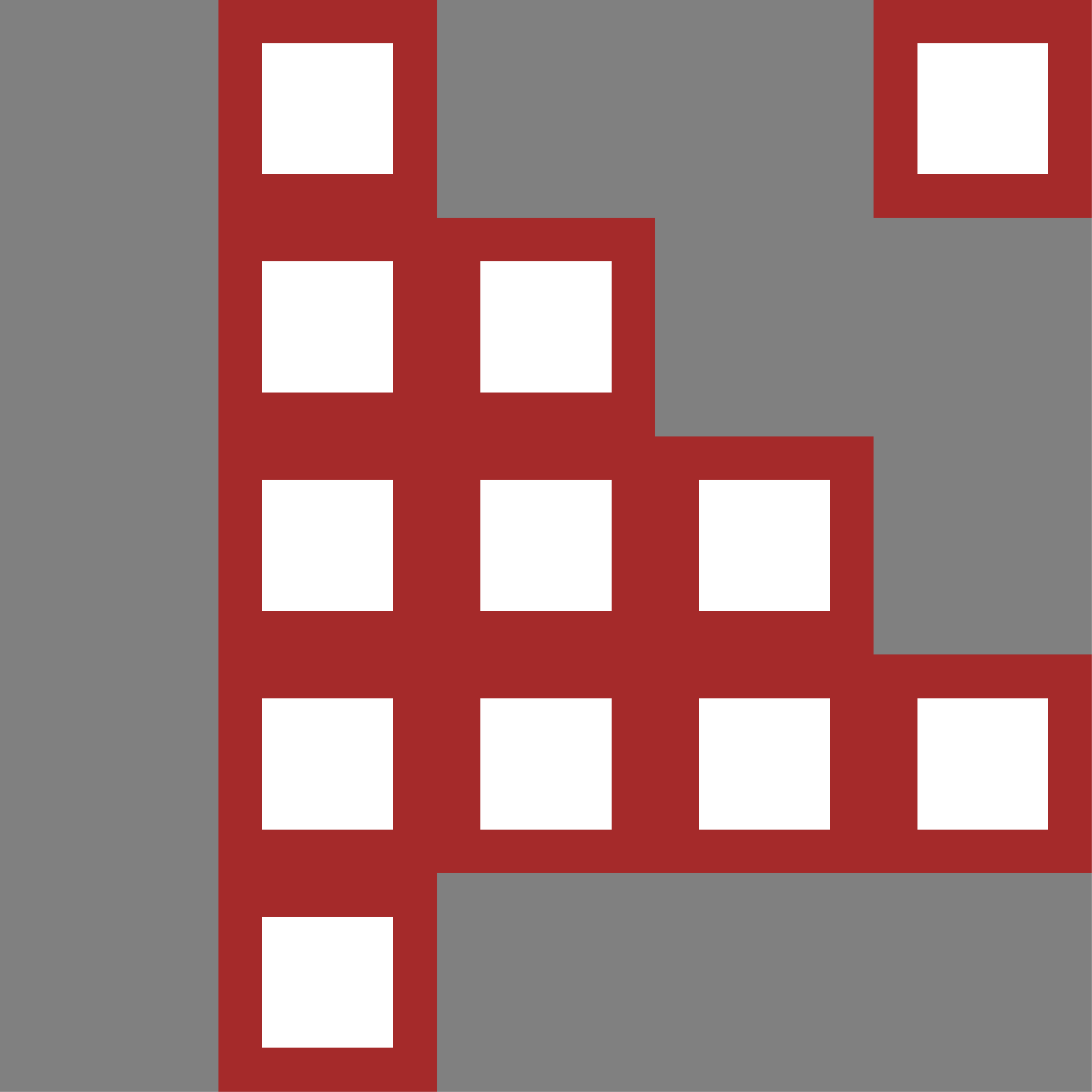}
        \caption{}
    \end{subfigure}
    \begin{subfigure}[t]{0.32\linewidth}
        \includegraphics[width=1\linewidth]{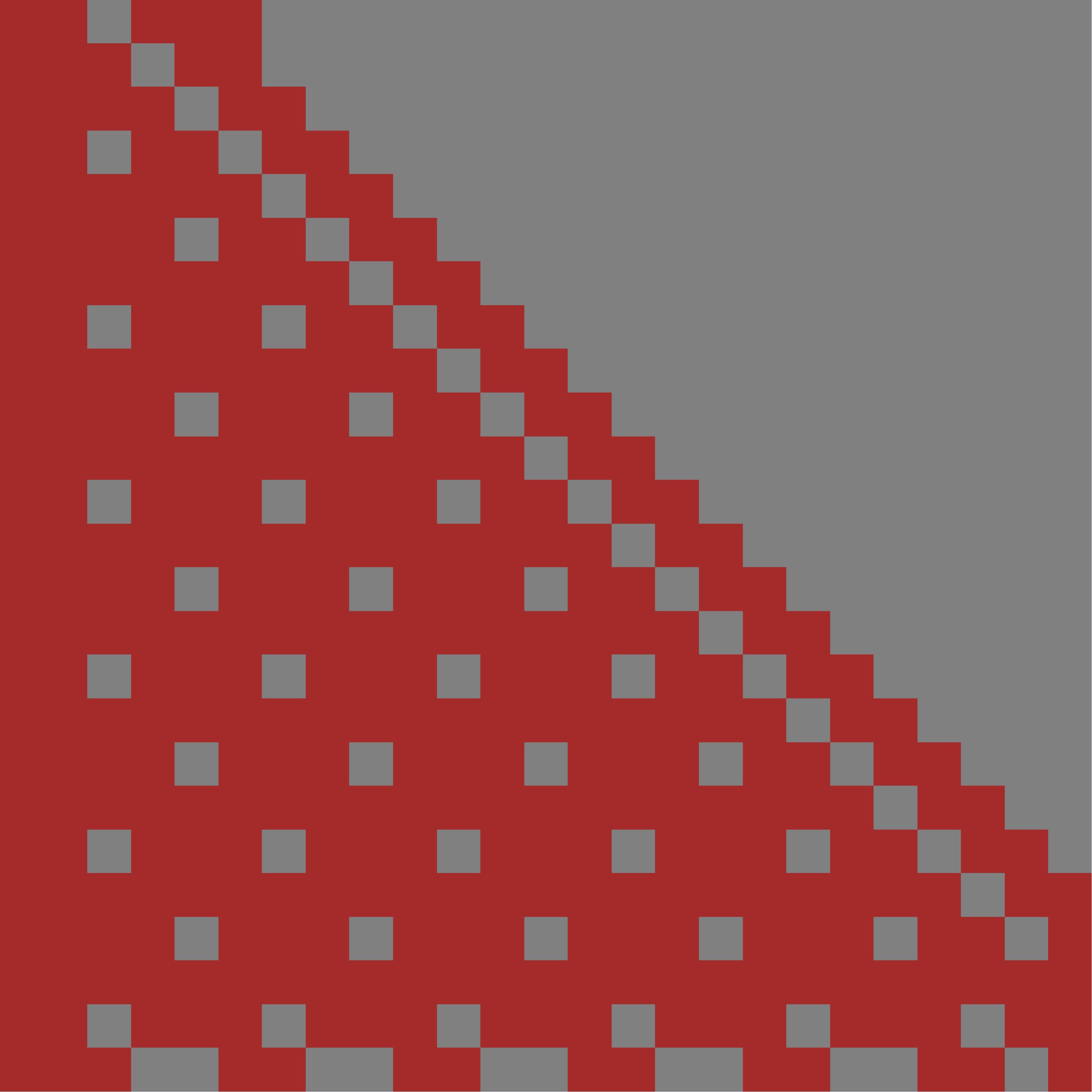}
        \caption{}
    \end{subfigure}
    \caption{}
    \label{fig:appx:compose_examples:random}
    \end{minipage}
    \caption*{The (a) desired, (b) composed and (c) flat levels for a (\autoref{fig:appx:compose_examples:hand}) hand-designed or (\autoref{fig:appx:compose_examples:random}) randomly generated target layout. Note that in the main text, we did not use the fitness results from the hand-designed layout and just illustrate it here for comparison.}
\end{figure*}

\subsection*{Composition vs flat}
\autoref{tab:appdx:compose_vs_flat} shows the hyperparameters we used when directly comparing \mcc and \pcgnn.

In this experiment, we have 20 random, but fixed, town layouts. Thus, for each method (\mcc and \pcgnn), we have the same desired levels.

\Autoref{fig:appx:compose_examples:hand,fig:appx:compose_examples:random} show the desired layout and the result from both the composed and flat methods. While the composed method does not generate perfect levels, they are significantly closer to the desired layouts compared to the flat generator. When using the flat method, the levels are very poor, not containing house structures at all.

\begin{table*}[h]
    \centering
    \caption{Hyperparameters used for the Composed vs Flat Towns}
    \label{tab:appdx:compose_vs_flat}
    \begin{adjustbox}{width=0.8\linewidth}
    \begin{tabular}{llll}
    \toprule
    Name & Flat (\pcgnn) & Composed Town & Composed House\\
    \midrule
    Level Size                                  & $25\times 25$     & $5\times 5$             & $5\times 5$\\
    One Hot Encoding                            & False             & False                   & False\\
    Generations                                 & 150               & 150                     & 150\\
    Population Size                             & 50                & 50                      & 50\\
    Number of Random Variables                  & 1                 & 1                       & 1\\
    Random Perturb Size                         & 0                 & 0                       & 0\\
    Number of Iterations                        & 5                & 5                      & 5\\
    Input Center Tile                           & Yes               & Yes                     & Yes\\
    Novelty                                     & No Novelty & No Novelty & No Novelty \\
    Level Starting Point                        & Default (Road)             & Default (Road)                   & Default (Air)\\
    \# of levels for fitness calculations       & 5                 & 5                       & 5\\
    \bottomrule
\end{tabular}
\end{adjustbox}
\end{table*}

\subsection*{Showcase}
\autoref{tab:appdx:showcase} details the hyperparameters used for each showcase result. The generator that we used to generate both towns and cities is denoted as ``Town \& City'', while the generator we used to illustrate the benefits of novelty is denoted as ``Novelty''.
We used the following fitness functions for these experiments:
\begin{LaTeXdescription}
    \item[House] The house was incentivised to be a hollow cube, with roof tiles at the top by setting the fitness to the average overlap between the generated house and the ``ideal'', hollow cube. Here we weighted novelty, intra-generator novelty and this fitness using the ratio $1:1:8$. The novelty distance function was the Hamming distance, which was denoted as \textit{Visual Diversity} by \citet{beukman2022Procedural}.
    \item[Garden] The garden had multiple fitness functions, incentivising it to have at least one tree and one flower, contain some water tiles (less than 5\%) and not be too dense (the fraction of grass tiles must be between $0.2$ and $0.7$ and trees should not be too close to each other). We weighted novelty, intra-generator novelty and this fitness using the ratio $1:1:4$. The novelty distance function was again the Hamming distance.
    \item[Town] This incentivised the distribution of tiles to be 40\% house, 30\% garden and 30\% road (by computing the square root of the Jensen-Shannon divergence between this distribution and the actual one), in addition to the houses being reachable via roads. This was calculated as $\frac{a + b}{2}$, with $a$ and $b$ defined in the \hyperref[sec:reach_fit]{Hyperparameters Section}.
    \item[City] The average of two fitnesses, the first one incentivising an equal number of houses, roads and gardens, calculated as $\frac{1}{3}\sum_{i \in \{ house, road, garden\}}\max(1 - 10 (n_i - \frac{1}{3})^2, 0)$ where $n_i$ is the fraction of tiles of type $i$. The second fitness is \textit{reachability}, again calculated as $\frac{a + b}{2}$.
    \item[Town \& City] This fitness incentivises the boundary of the town to be houses, with an inner ring of roads, and a large interior section of gardens. This fitness was calculated as the average overlap between the ideal layout and the actual level for each section. This generator also used a context size of $2$, i.e. it took in a $5\times 5$ block around the current tile, instead of the normal $3\times 3$.
    \item[Novelty] This used the exact same fitness functions as the ``Town'' generator. It additionally used intra-generator novelty, and normal novelty, using the $2\times 2$ tile-based KL-Divergence~\citep{lucas2019Tile} as a distance metric. This is discussed more in \refappendixQuantNov{sec:appdx:novresults}.
\end{LaTeXdescription}
\begin{table*}
    \centering
    \caption{This table details the hyperparameters used for the showcase images shown in the main text.}
    \label{tab:appdx:showcase}
    \begin{adjustbox}{width=1\linewidth}
    \begin{tabular}{lllllll}
    \toprule
    Name                                   & House & Garden & Town & City  & Town  \& City                                              & Novelty       \\
    \midrule
    One Hot Encoding                       & False            & False           & False            & False             & False          & False         \\
    Generations                            & 500              & 100             & 150              & 150               & 350            & 150                \\
    Population Size                        & 20               & 20              & 50               & 50                & 30             & 50                \\

    Number of Random Variables             & 1                & 2               & 1                & 1                 & 1              & 2                 \\
    Random Perturb Size                    & 0                & 0.1565          & 0                & 0                 & 0.1            & 0.2                \\
    Number of Iterations                   & 10               & 1               & 10               & 10                & 5              & 10                \\
    Input Center Tile                      & Yes              & No              & Yes              & Yes               & Yes            & Yes                \\
    Level Starting Point                   & Random           & Random          & Default (Road)          & Default (Road) & Random     & Random                \\
    \# of levels for fitness calculations  & 15               & 15              & 5                & 5                & 15              & 10                \\
    Novelty Distance Function                & Hamming          & Hamming         & No Novelty       & No Novelty        & No Novelty   & Various                \\
    \bottomrule
\end{tabular}
\end{adjustbox}
\end{table*}

\section{Coalescing Comparison}
\label{sec:appdx:coalesce}
\begin{figure*}[h!]
  \centering
  \begin{subfigure}{0.45\linewidth}
    \includegraphics[width=1\linewidth]{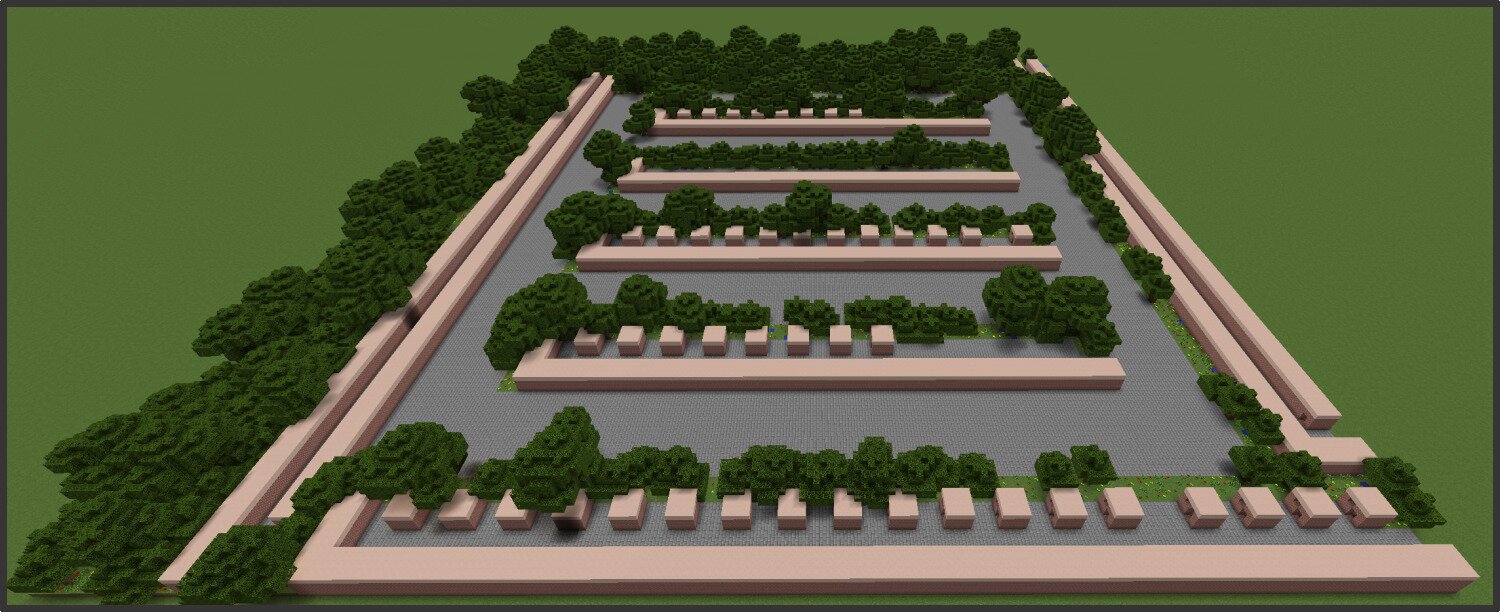}
      \caption{}
      \label{fig:mcc:showcase:town1}
  \end{subfigure}
  \begin{subfigure}{0.45\linewidth}
      \includegraphics[width=1\linewidth]{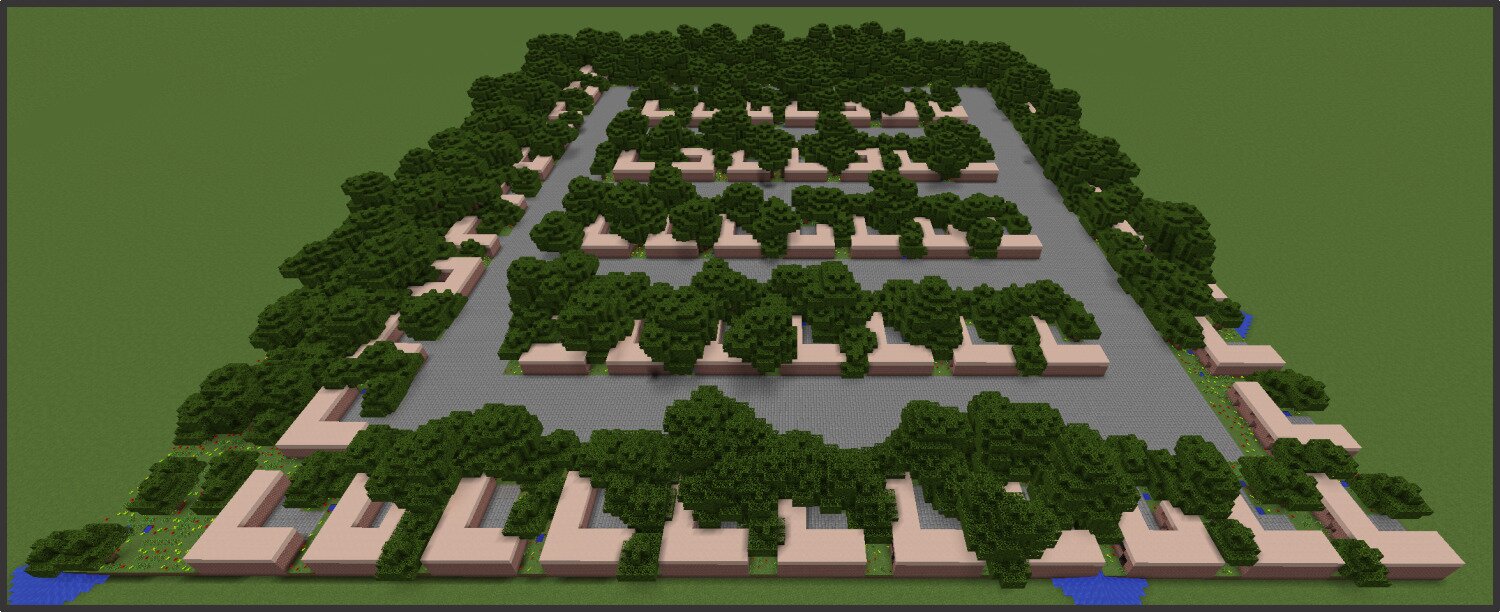}
      \caption{}
      \label{fig:mcc:showcase:town2}
  \end{subfigure}
  \caption{Demonstrating some generated cities. In (a) and (b) we use the same generators, but (a) uses coalescing whereas (b) does not.}
    \label{fig:mcc:showcase:appendix:coalesce_vs_not}
\end{figure*}

Here we briefly provide an additional qualitative demonstration; in particular, illustrating the difference between using coalescing and not. The generators used are the same as the annotated example in the main text. 
These results are shown in \autoref{fig:mcc:showcase:appendix:coalesce_vs_not}, where we use the same generators, but (a) uses coalescing whereas (b) does not. In \autoref{fig:mcc:showcase:town1}, we can see that there are fewer, but larger and more rectangular towns. In \autoref{fig:mcc:showcase:town2}, we do not coalesce, and each town is the same size.

\section{Additional Novelty Results}
\label{sec:appdx:novresults}
In this section, we illustrate quantitative diversity scores. In particular, we use a town level with hyperparameters listed in \autoref{tab:appdx:showcase}, under the Novelty column.
Each setting uses at least two fitness functions, reachability and probability, weighed equally. Three of the settings also use two additional fitness functions, intra-generator novelty and standard novelty-search, with weights:
$\text{Intra-Novelty}: \text{Novelty}: \text{Reachability}: \text{Probability} = 1:1:4:4$. We weigh novelty less than the feasibility fitness functions, as it is generally easier to generate diverse levels compared to generating feasible ones (e.g., a random generator will have a high diversity and low feasibility). If the novelty distance function is weighted too much, we found that the generators overprioritised novelty.
The aspect we change in this section is simply the novelty distance function we use, and we consider the following ones:
\begin{LaTeXdescription}
    \item[None] This uses no novelty objectives.
    \item[Hamming] This uses the Visual Diversity (also called Hamming distance), where the distance between two levels is how many tiles are different between them.
    \item[PHash] This uses the simple perceptual image hash~\citep{monga2006Perceptual,hadmi2012Perceptual} as implemented using the ImageHash library.\footnote{\url{https://github.com/JohannesBuchner/imagehash}}
    \item[KL$(2\times 2)$] This uses the tile-based KL-Divergence between two levels~\citep{lucas2019Tile}. In particular, for each level, it builds up a probability distribution over $2\times 2$ tile patterns, and then computes the KL-Divergence between the two distributions. Since the KL-Divergence is not symmetric, we set this distance to $d(A, B) = \frac{KL(A||B) + KL(B||A)}{2}$. Furthermore, since the KL-Divergence is unbounded (and \pcgnn requires normalised fitness functions), we normalise the distance as $d(A, B) = clip(\frac{d(A, B)}{8}, 0, 1)$, as this resulted in similar novelty fitness values as the other, normalised metrics.
\end{LaTeXdescription}
We evaluate these different metrics as follows:
\begin{enumerate}
    \item For each seed and at each generation, take the generator with the highest fitness value.
    \item Generate $100$ $10\times 10$ levels using this generator.
    \item Compute the average pairwise distance between these levels, using the Hamming, PHash, KL$(2\times 2)$ and KL$(3\times 3)$ metrics.
\end{enumerate}

We follow this procedure because the generator with the highest fitness is, by default, the generator that will be used when generating levels. Secondly, we wish to measure how diverse the levels are when generating several from a single generator. 
Since there is a large amount of noise, for clarity, we additionally smooth the plots using a Savitzky-Golay filter~\citep{savitzky1964Smoothing} using a window size of $11$ and a polynomial order of $3$.

\autoref{fig:appdx:nov:all} shows the results. In general, we can see that having a novelty objective increases the diversity of the generators. Using the Hamming distance during training results in the largest amount of diversity across all four metrics, followed by PHash and KL$(2\times 2)$. As expected, not incentivising novelty results in the lowest diversity.
For these plots, we simply show the diversity of single $10\times 10$ town levels instead of measuring the diversity of composed levels. The reason is that many of these metrics are local; for instance, the KL-Divergence only measures the similarity of small $2\times 2$ tile patterns. Thus, the diversity is largely controlled by the lower-level generators. We therefore opt to qualitatively analyse the compositional effects of novelty in the main text.

In \autoref{fig:appdx:novelty_show:all}, we show some generated levels using each of the distance functions. In general, the Hamming distance results in levels with more randomness, while the other distance functions result in more geometric shapes, while still generating diverse levels.
\newcommand{\awww}{0.49}
\newcommand{\awwww}{0.8}
\begin{figure*}
    \centering
    \begin{subfigure}{\awww\linewidth}
        \includegraphics[width=1\linewidth]{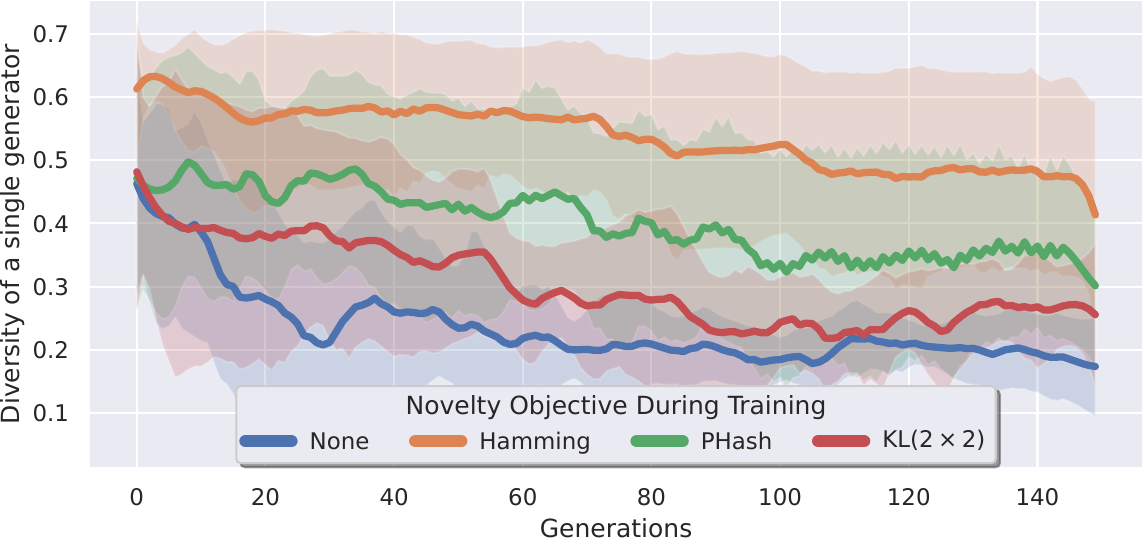}
        \caption{Hamming}\label{fig:appdx:nov:hamming}
    \end{subfigure}\hfill
    \begin{subfigure}{\awww\linewidth}
        \includegraphics[width=1\linewidth]{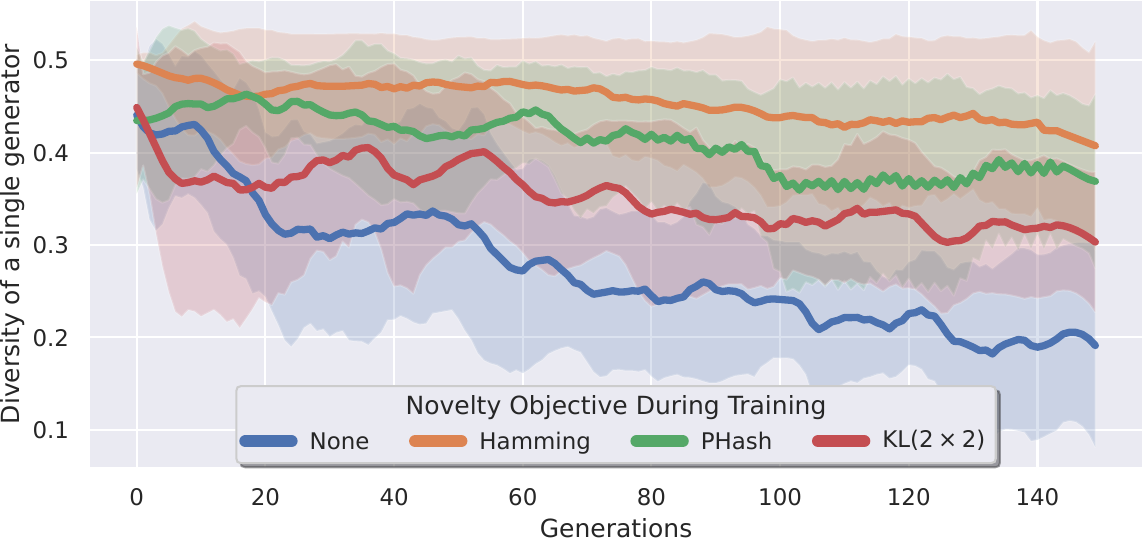}
        \caption{PHash}\label{fig:appdx:nov:phash}
    \end{subfigure}%

    \begin{subfigure}{\awww\linewidth}
        \includegraphics[width=1\linewidth]{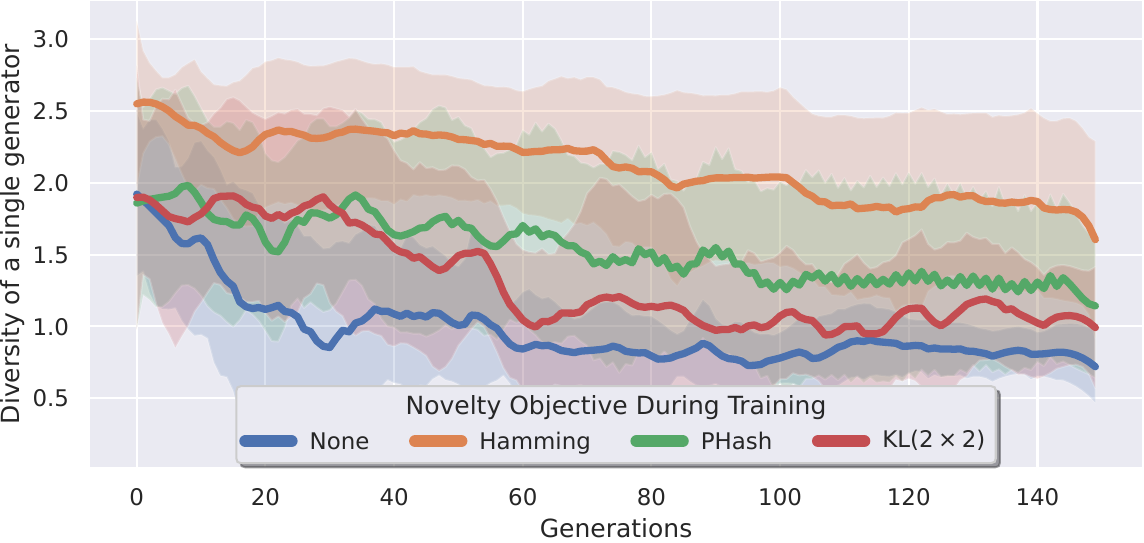}
        \caption{$\text{KL}(2\times 2)$}\label{fig:appdx:nov:kl2}
    \end{subfigure}\hfill
    \begin{subfigure}{\awww\linewidth}
        \includegraphics[width=1\linewidth]{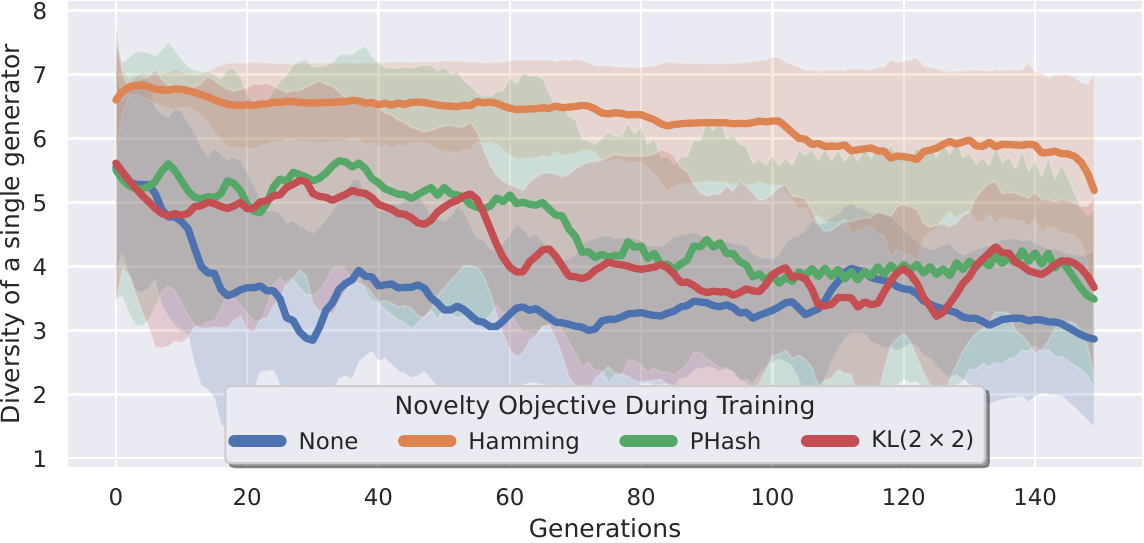}
        \caption{$\text{KL}(3\times 3)$}\label{fig:appdx:nov:kl3}
    \end{subfigure}
    \caption{The average pairwise diversity over $100$ generated levels from the generator with the highest fitness over time. Each plot contains the results when using a particular metric, and each line indicates the novelty objective used during training.}
    \label{fig:appdx:nov:all}
\end{figure*}

{
\setlength{\fboxrule}{2pt}%
\setlength{\fboxsep}{0pt}%
\begin{figure*}
    \centering
    \begin{subfigure}{\awwww\linewidth}
        \fbox{\includegraphics[width=0.48\linewidth]{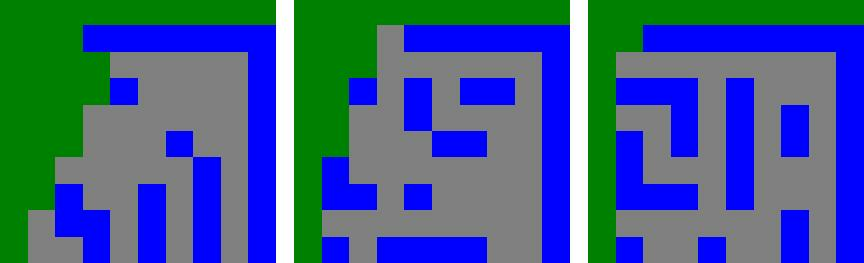}}\hfill
        \fbox{\includegraphics[width=0.48\linewidth]{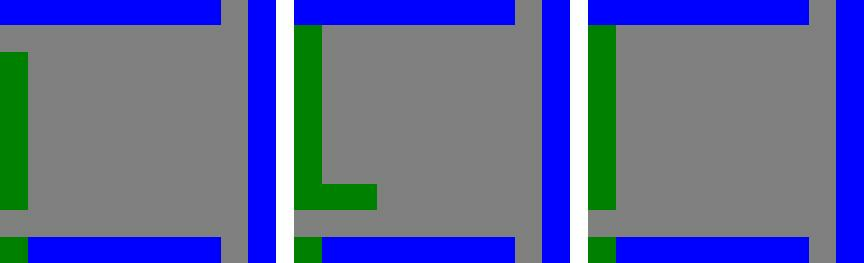}}
        \caption{No Novelty}\label{fig:appdx:novelty_show:none}
    \end{subfigure}

    \begin{subfigure}{\awwww\linewidth}
        \fbox{\includegraphics[width=0.48\linewidth]{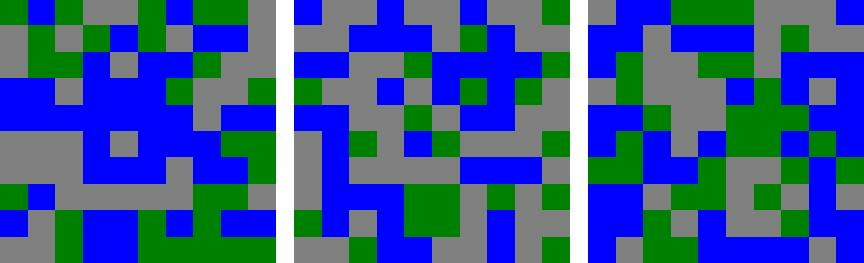}}\hfill
        \fbox{\includegraphics[width=0.48\linewidth]{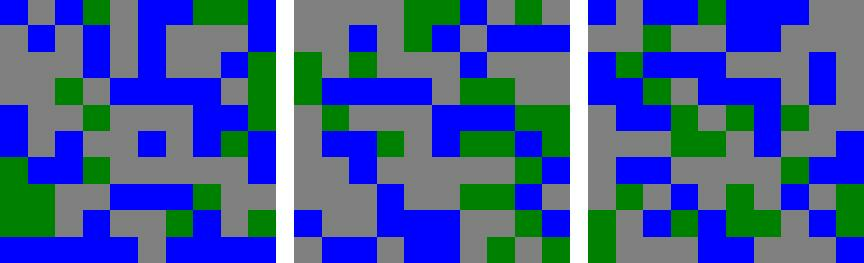}}
        \caption{Hamming}\label{fig:appdx:novelty_show:hamming}
    \end{subfigure}

    \begin{subfigure}{\awwww\linewidth}
        \fbox{\includegraphics[width=0.48\linewidth]{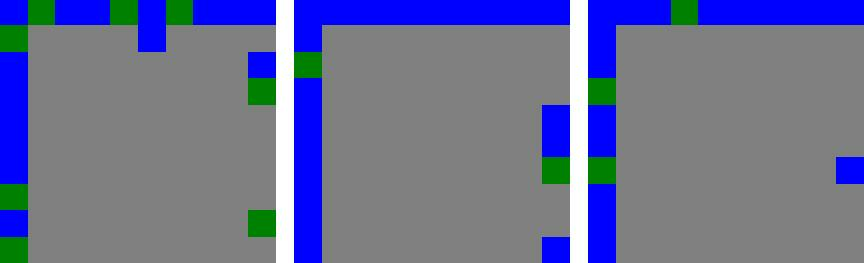}}\hfill
        \fbox{\includegraphics[width=0.48\linewidth]{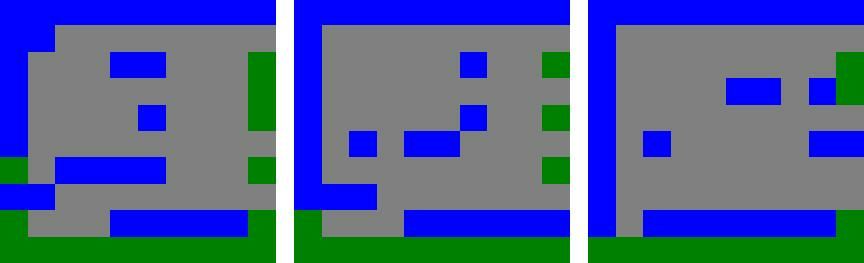}}
        \caption{PHash}\label{fig:appdx:novelty_show:phash}
    \end{subfigure}

    \begin{subfigure}{\awwww\linewidth}
        \fbox{\includegraphics[width=0.48\linewidth]{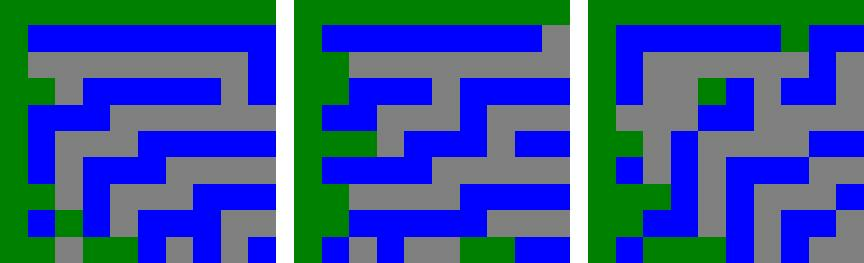}}\hfill
        \fbox{\includegraphics[width=0.48\linewidth]{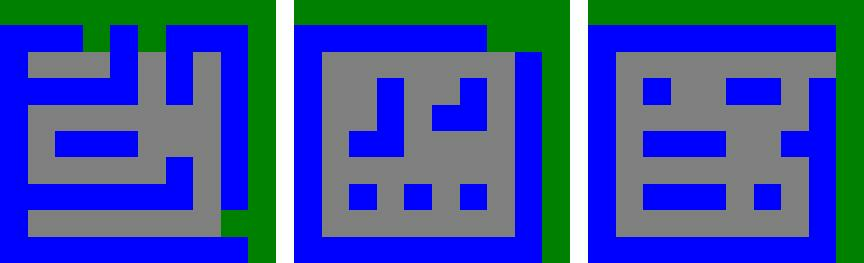}}
        \caption{$\text{KL}(2\times 2)$}\label{fig:appdx:novelty_show:kl2}
    \end{subfigure}
    \caption{Showing three images from two generators (left and right) from the same experiment and different random seeds, when using (a) no novelty and (b,c,d) the three other distance functions during training. Blue tiles are houses, green are gardens and grey tiles are roads.}
    \label{fig:appdx:novelty_show:all}
\end{figure*}

}
\end{document}